\pdfoutput=1

\documentclass[11pt]{article}

\usepackage[]{acl}

\usepackage{times}
\usepackage{latexsym}

\usepackage[T1]{fontenc}

\usepackage[utf8]{inputenc}
\usepackage{textcomp}
\DeclareUnicodeCharacter{266B}{\textmusicalnote}
\usepackage{microtype}

\usepackage{inconsolata}

\usepackage{multicol}
\usepackage{booktabs}
\usepackage{makecell}
\usepackage{amsmath, amssymb}
\usepackage{multirow}
\usepackage{mathtools}
\usepackage{xcolor}
\usepackage{enumitem}
\usepackage{float}
\usepackage{graphicx}
\usepackage{upgreek}
\usepackage{seqsplit}
\usepackage{color,soul}
\usepackage{arydshln}
\usepackage{placeins}
\usepackage{pifont}
\usepackage{amssymb}
\usepackage{bbm}

\usepackage{amsmath,amsfonts,bm}

\def\eqref#1{equation~\ref{#1}}

\def\1{\bm{1}}

\def\va{{\bm{a}}}

\def\vc{{\bm{c}}}
\def\vd{{\bm{d}}}

\def\vq{{\bm{q}}}

\def\vx{{\bm{x}}}
\def\vy{{\bm{y}}}

\DeclareMathAlphabet{\mathsfit}{\encodingdefault}{\sfdefault}{m}{sl}
\SetMathAlphabet{\mathsfit}{bold}{\encodingdefault}{\sfdefault}{bx}{n}

\usepackage{caption}
\usepackage{subcaption}
\title{Adaptive-RAG: Learning to Adapt Retrieval-Augmented \\ Large Language Models through Question Complexity}

\author{Soyeong Jeong$^1$
        \quad Jinheon Baek$^2$
        \quad Sukmin Cho$^1$
        \quad Sung Ju Hwang$^{1, 2}$
        \quad Jong C. Park$^1$\thanks{\hspace{0.2cm}Corresponding author} \\
        School of Computing$^1$ \quad Graduate School of AI$^2$ \\
        Korea Advanced Institute of Science and Technology$^1$$^,$$^2$\\
       \texttt{\{starsuzi,jinheon.baek,nelllpic,sjhwang82,jongpark\}@kaist.ac.kr}}

\begin{document}
\maketitle

\begin{abstract}

Retrieval-Augmented Large Language Models (LLMs), which incorporate the non-parametric knowledge from external knowledge bases into LLMs, have emerged as a promising approach to enhancing response accuracy in several tasks, such as Question-Answering (QA). However, even though there are various approaches dealing with queries of different complexities, they either handle simple queries with unnecessary computational overhead or fail to adequately address complex multi-step queries; yet, not all user requests fall into only one of the simple or complex categories. In this work, we propose a novel adaptive QA framework that can dynamically select the most suitable strategy for (retrieval-augmented) LLMs from the simplest to the most sophisticated ones based on the query complexity. Also, this selection process is operationalized with a classifier, which is a smaller LM trained to predict the complexity level of incoming queries with automatically collected labels, obtained from actual predicted outcomes of models and inherent inductive biases in datasets. This approach offers a balanced strategy, seamlessly adapting between the iterative and single-step retrieval-augmented LLMs, as well as the no-retrieval methods, in response to a range of query complexities. We validate our model on a set of open-domain QA datasets, covering multiple query complexities, and show that ours enhances the overall efficiency and accuracy of QA systems, compared to relevant baselines including the adaptive retrieval approaches. Code is available at: \url{https://github.com/starsuzi/Adaptive-RAG}.

\end{abstract}
\section{Introduction}
\begin{figure}
    \centering
    \includegraphics[width=0.9\linewidth]{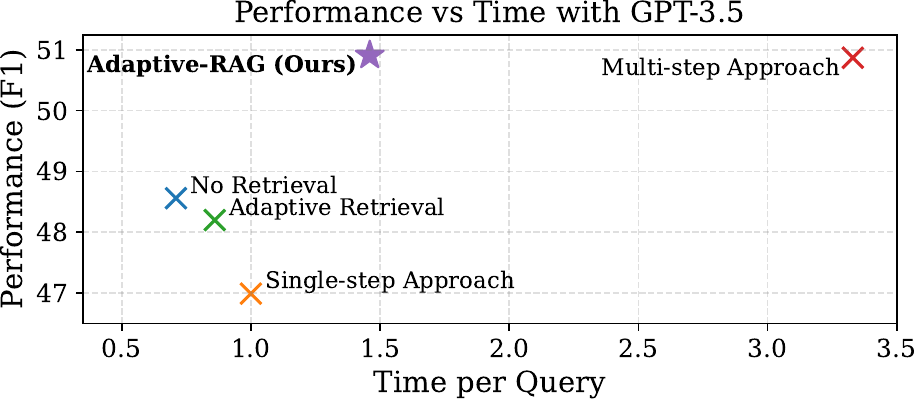}
    \vspace{-0.110in}
    \caption{QA performance (F1) and efficiency (Time/Query) for different retrieval-augmented generation approaches. We use the GPT-3.5-Turbo-Instruct as the base LLM.}
    \label{fig:perf_effi_gpt}
    \vspace{-0.225in}
\end{figure}

Recent Large Language Models (LLMs)~\cite{GPT-3, GPT-4, LLaMA2, PaLM2} have shown overwhelming performances across diverse tasks, including question-answering (QA)~\cite{hotpotqa, nq}. However, they still generate factually incorrect answers since their knowledge solely relies on their parametric memory~\cite{RQA, adaptiveRetrieval}. Meanwhile, memorizing all the (ever-changing) world knowledge may not be possible. To address this problem, retrieval-augmented LLMs~\cite{retro, atlas, replug}, which incorporate non-parametric knowledge into LLMs with additional retrieval modules, have gained much increasing attention. Specifically, these models access a knowledge base, which serves as an extensive repository of information across various subjects and disciplines, to retrieve information relevant to the given input, and then incorporate the retrieved information into LLMs, which enables them to stay accurate and current with the world knowledge.

\begin{figure*}
    \centering
    \includegraphics[width=0.99\linewidth]{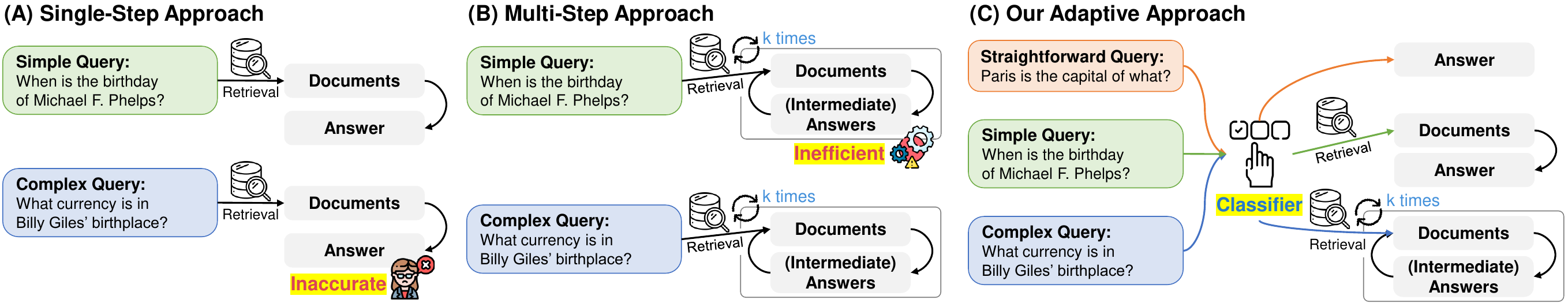}
    \vspace{-0.1in}
    \caption{A conceptual comparison of different retrieval-augmented LLM approaches to question answering. (A) In response to a query, this single-step approach retrieves relevant documents and then generates an answer. However, it may not be sufficient for complex queries that require multi-step reasoning. (B) This multi-step approach iteratively retrieves documents and generates intermediate answers, which is powerful yet largely inefficient for the simple query since it requires multiple accesses to both LLMs and retrievers. (C) Our adaptive approach can select the most suitable strategy for retrieval-augmented LLMs, ranging from iterative, to single, to even no retrieval approaches, based on the complexity of given queries determined by our classifier.}
    \label{fig:concept}
    \vspace{-0.175in}
\end{figure*}

A particularly salient application of retrieval-augmented LLMs is to handling QA tasks, whose goal is to provide correct answers in response to user queries, especially those of high complexity. Early work on retrieval-augmented LLMs focuses primarily on single-hop queries~\cite{internet-QA, ram2023ralm}, whose answers are typically found within a single document; therefore, this approach involves retrieving a relevant document based on the query and subsequently integrating this information into QA models to formulate a response. However, unlike this single-hop QA, some queries require connecting and aggregating multiple documents, which are, furthermore, often not answerable through a single-step process of retrieval-and-response. An example query is `When did the people who captured Malakoff come to the region where Philipsburg is located?', which requires four reasoning steps to solve. Therefore, to effectively handle such complex queries, recent studies have concentrated largely on multi-step and multi-reasoning QA, which requires iterative accesses to both LLMs and retrievers multiple times~\cite{self-ask, ircot}, at the cost of heavy computational overheads.

Yet, we should rethink: In a real-world scenario, are all the requests from users complex? Instead, users might often ask simple and straightforward questions, while only occasionally asking complex ones. Specifically, a query such as `Paris is the capital of what?' is likely to be asked more frequently, compared to the aforementioned multi-step query, and this simpler query might also be easily answered by the LLMs themselves, without accessing external knowledge. In other words, a multi-step QA approach could give rise to unnecessary computational overhead for simple queries, even though it would be vital for complex queries (see Figure~\ref{fig:concept} (A)). On the other hand, handling complex queries with single-step-retrieval or even non-retrieval strategies would be largely insufficient (Figure~\ref{fig:concept} (B)). This suggests the need for an adaptive QA system, which can dynamically adjust the operational strategies of retrieval-augmented LLMs based on the query complexity. While some recent approaches are capable of doing this based on the frequency of entities in queries~\cite{adaptiveRetrieval} or on the generated outputs from models for multi-step QA~\cite{ircot}, they are still suboptimal: the former methods are overly simplistic, failing to consider multi-hop queries; meanwhile, the latter are excessively complex, terminating answer solving steps after several rounds of module access.

In this work, considering diverse complexity levels of real-world queries, we argue that previous one-size-fits-all approaches might be inadequate to cover all of them. Instead, we propose to select the most suitable strategy from a range of (retrieval-augmented) LLMs, each of which is tailored to the specific complexity of the input query. Notably, a critical step in this process is pre-defining the query complexity, which is instrumental in determining the most fitting model to it. In this work, we operationalize this process with a novel classifier, which is a smaller model trained to predict the complexity level of incoming queries (see Figure~\ref{fig:concept} (c)). Moreover, we automatically collect its training datasets without human labeling, by leveraging the predicted outcomes (i.e., which models accurately respond to which queries) as well as by capitalizing on the inherent biases in existing datasets (i.e., samples in the datasets are designed either for single-step or for multi-step QA scenarios). This proposed method can offer a robust middle ground among the iterative LLM augmentation methods for complex queries, single-step methods for simpler queries, and even no-retrieval-augmented methods for the most straightforward queries (answerable by LLMs themselves), thus significantly enhancing the overall efficiency and accuracy, as shown in Figure~\ref{fig:perf_effi_gpt}. We refer to our framework as Adaptive Retrieval-Augmented Generation (Adaptive-RAG).

We validate Adaptive-RAG using benchmark open-domain QA datasets, covering a wide range of query complexity from single-hop~\cite{squad, tqa, nq} to multi-hop~\cite{hotpotqa, 2hop, trivedi-etal-2022-musique} queries. The experimental results show that ours significantly improves the overall accuracy and efficiency, compared to the prior adaptive strategies, on multiple LLMs, such as GPT-3.5~\cite{GPT-3} and FLAN-T5 series~\cite{flan}.

Our contributions and findings are threefold:

\vspace{-0.075in}

\begin{itemize}[itemsep=0.3mm, parsep=1pt, leftmargin=*]
  \item We point out the realistic scenario of queries of varying complexities, and find out that existing retrieval-augmented generation approaches tend to be overly simple or complex.
  \item We adapt retrieval-augmented LLMs to the query complexity assessed by the classifier, which enables the utilization of the most suitable approach tailored to each query.
  \item We show that our Adaptive-RAG is highly effective and efficient, balancing between the complexity and the simplicity for diverse queries.
\end{itemize}

\section{Related Work}

\paragraph{Open-domain QA}
Open-domain QA is the task of accurately answering a query by sourcing for query-relevant documents, and then interpreting them to provide answers~\cite{DBLP:conf/acl/ChenFWB17, ODQA/survey}, which, thus, generally involves two modules: a retriever~\cite{DPR, ANCE} and a reader~\cite{BERTserini, FiD, DBLP:conf/emnlp/JeongBCHP23}. 
Along with the emergence of LLMs with superior reasoning capabilities thanks to their billion-sized parameters~\cite{EmergentLLMs}, a synergy between LLMs and retrievers has led to significant advancements~\cite{internet-QA, ram2023ralm}. Specifically, this integration has been shown to enhance Open-domain QA by mitigating the hallucination problem from LLMs through strengthened reasoning abilities of the reader, as well as utilizing the retrieved, external documents~\cite{DBLP:conf/emnlp/ChoSJP23}. Despite these advancements for single-hop retrieval-augmented LLMs, however, the complexity of some queries needs a more complex strategy.

\paragraph{Multi-hop QA}
Multi-hop QA is an extension of conventional Open-domain QA, which additionally requires the system to comprehensively gather and contextualize information from multiple documents (often iteratively), to answer more complex queries~\cite{musique, hotpotqa}. In the realm of multi-hop QA, the approach to iteratively access both LLMs and the retrieval module is generally employed. Specifically,~\citet{dsp},~\citet{self-ask},~\citet{DBLP:conf/ecir/PereiraFLN23} and~\citet{decomp} proposed to first decompose the multi-hop queries into simpler single-hop queries, repeatedly access the LLMs and retriever to solve these sub-queries, and merge their solutions to formulate a complete answer. In contrast to this decomposition-based approach, other recent studies, such as~\citet{react} and~\citet{ircot}, explored the interleaving of Chain-of-Thought reasoning~\cite{cot} — a method where a logical sequence of thoughts is generated — with document retrieval, repeatedly applying this process until the reasoning chain generates the answer. In addition,~\citet{ActiveRetrieval} introduced an approach to repeatedly retrieving new documents if the tokens within generated sentences have low confidence. However, the aforementioned methods overlooked the fact that, in real-world scenarios, queries are of a wide variety of complexities. Therefore, it would be largely inefficient to iteratively access LLMs and retrievers for every query, which might be simple enough with a single retrieval step or even only with an LLM itself. 

\paragraph{Adaptive Retrieval}
To handle queries of varying complexities, the adaptive retrieval strategy aims to dynamically decide whether to retrieve documents or not, based on each query's complexity.
In this vein,~\citet{adaptiveRetrieval} proposed to decide the query's complexity level based on the frequency of its entities and suggested using the retrieval modules only when the frequency falls below a certain threshold. However, this approach, focusing solely on the binary decision of whether to retrieve or not, may not be sufficient for more complex queries that require multiple reasoning steps.
Additionally, \citet{DBLP:conf/emnlp/0003LSM21} proposed an approach that performs a fixed set of operations (retrieving, reading, and reranking) multiple times until the answer is derived for the given query, which is built upon traditional BERT-like LMs. However, unlike our Adaptive-RAG which pre-determines the query complexity and adapts the operational behavior of any off-the-shelf LLMs accordingly, this approach applies the same fixed operations to every query regardless of its complexity but also necessitates additional specific training to LMs. Concurrent to our work,~\citet{self-rag} suggested training a sophisticated model to dynamically retrieve, critique, and generate the text. Nevertheless, we argue that all the aforementioned adaptive retrieval methods that rely on a single model might be suboptimal in handling a variety of queries of a range of different complexities since they tend to be either overly simple or complex for all the input queries, which demands a new approach that can select the most suitable strategy of retrieval-augmented LLMs tailored to the query complexity.

\section{Method}
In this section, we describe our approach to adapting retrieval-augmented LLMs, by pre-determining the query complexity and then selecting the most fitting strategies for retrieval-augmented LLMs.

\subsection{Preliminaries}
\label{method:pre}
We begin with preliminaries, formally introducing different strategies of retrieval-augmented LLMs.

\paragraph{Non Retrieval for QA}
Let us first define an LLM as a model $\texttt{LLM}$, which takes a sequence of tokens $\vx = [x_1, x_2, ..., x_n]$ as an input and then generates a sequence of tokens $\vy = [y_1, y_2, ..., y_n]$ as an output, which is formalized as follows: $\vy = \texttt{LLM}(\vx)$. Then, in our problem setup for QA, $\vx$ and $\vy$ become the input query ($\vq$) from the user and the generated answer ($\bar{\va}$) from the LLM, respectively: $\vq = \vx$ and $\bar{\va} = \vy$. Also, subsequently, the most naïve LLM-powered QA model can be represented as follows: $\bar{\va} = \texttt{LLM}(\vq)$. Ideally, $\bar{\va}$ should match the actual correct answer $\va$. This non-retrieval-based QA method is highly efficient and could be a somewhat promising approach to handling easy queries, as the size of LLMs becomes extremely large with its effect on storing a large amount of knowledge. However, this approach is largely problematic on queries that require precise or concurrent knowledge of specific people, events, or any subjects beyond the LLMs' internal knowledge.

\paragraph{Single-step Approach for QA}
To address the aforementioned scenarios where $\texttt{LLM}$ may struggle with queries that are not answerable by $\texttt{LLM}$ itself, we can utilize the external knowledge $\vd$, which includes useful information for queries, retrieved from the external knowledge source $\mathcal{D}$ that could be an encyclopedia (e.g., Wikipedia) consisting of millions of documents. Specifically, to obtain such $\vd$ from $\mathcal{D}$, a specific retrieval model is necessary, which returns documents based on their relevance with the given query. This process can be formulated as follows: $\vd = \texttt{Retriever}(\vq; D)$, where $\texttt{Retriever}$ is the retrieval model, with $\vd \in \mathcal{D}$. Here, we can use any off-the-shelf retriever~\cite{bm25, DPR}. 

After the retrieval step is done, we now have a pair of query $\vq$ and its relevant documents $\vd$. Then, in order to augment LLMs with this retrieved external knowledge, we can incorporate it into the input of LLMs, represented as follows: $\bar{\va} = \texttt{LLM}(\vq, \vd)$. This process allows LLMs to gain access to external information contained in $\vd$, which can provide the supplementary context that the internal knowledge of $\texttt{LLM}$ lacks, which can subsequently improve the accuracy and concurrency of LLMs for QA. 

\paragraph{Multi-step Approach for QA}
Even though the aforementioned single-step approach offers significant improvements over non-retrieval for $\vq$ that requires external knowledge, it encounters notable limitations, particularly when dealing with complex queries that necessitate synthesizing information from multiple source documents and reasoning over them. This is where a multi-step approach and reasoning for QA become essential. 

In this multi-step approach, $\texttt{LLM}$ interacts with \texttt{Retriever} in several rounds, progressively refining its understanding of $\vq$, until it formulates the final answer from findings accumulated across these multiple steps. Specifically, the process begins with the initial query $\vq$, and at every retrieval step $i$, new documents $\vd_i$ are retrieved from $\mathcal{D}$ and then incorporated into the input of LLMs, as follows: $\bar{\va}_i = \texttt{LLM}(\vq, \vd_i, \vc_i)$, where the additional context $\vc_i$ can be composed of previous documents and outcomes $(\vd_1, \vd_2, ..., \vd_{i-1}, \bar{\va}_1, \bar{\va}_2, ..., \bar{\va}_{i-1})$, and $\vd_i = \texttt{Retriever}(\vq, \vc_i; D)$\footnote{It is worth noting that implementations of the LLM and retriever vary across different multi-step retrieval-augmented LLM approaches~\cite{ircot, self-ask, react}; therefore, the context $\vc_i$ may incorporate none, some, or all of the previous documents and answers.}. We would like to note that this iterative, multi-step process enables $\texttt{LLM}$ to construct a more comprehensive and extensive foundation to solve queries effectively, specifically adept at complex multi-hop queries where answers depend on interconnected pieces of information. However, it is important to recognize that this multi-step approach can be resource-intensive due to the repeated accesses to $\texttt{Retriever}$ and $\texttt{LLM}$, which entail substantial computational costs.

\subsection{Adaptive-RAG: Adaptive Retrieval-Augmented Generation}
\label{method:ours}

We now introduce our adaptive retrieval-augmented LLMs, which are built upon three different strategies described in the previous section, and which are designed to select the most suitable strategy according to the complexity of queries.

\paragraph{Adapting Retrieval-Augmented LLMs}
Note that in real-world scenarios, not all $\vq$ from users have the same level of complexity, necessitating tailored strategies for handling each query. In other words, employing the most basic, non-retrieval-based approach $\texttt{LLM}(\vq)$ to respond to the complex query $\vq$ would be also ineffective (Figure~\ref{fig:concept}, A); conversely, using a more elaborate multi-step approach $ \texttt{LLM}(\vq, \vd, \vc)$ for simple $\vq$ would be inefficient (Figure~\ref{fig:concept}, B). Therefore, our adaptive framework is designed to dynamically adjust the query-handling strategy of retrieval-augmented LLMs, which is achieved by determining the complexity of each query before attempting a solution. Notably, this framework can offer a robust middle ground with a range of solutions, from the simplest approach for the most straightforward queries, to the one-step approach for moderate queries, and up to the most comprehensive and rigorous approach for complex queries. In addition, since the operations of $\texttt{LLM}$ and $\texttt{Retriever}$ remain consistent regardless of inputs to them, our method can seeminglessly go back and forth across queries of different complexities, without changing the internal model architecture or parameters during adaption.

\paragraph{Query Complexity Assessment}
To operationalize our adaptive retrieval-augmented LLM framework, we should determine the query complexity, and to achieve this, we propose to model a complexity classifier, whose goal is to return the appropriate complexity level of the given query. Specifically, given the query $\vq$, our classifier can be formulated as follows: $o = \texttt{Classifier}(\vq)$, where $\texttt{Classifier}$ is a smaller Language Model that is trained to classify one of three different complexity levels and $o$ is its corresponding class label. In our classifier design, there are three class labels: `A', `B', and `C', where `A' indicates that $\vq$ is straightforward and answerable by $\texttt{LLM}(\vq)$ itself, `B' indicates that $\vq$ has the moderate complexity where at least a single-step approach $\texttt{LLM}(\vq, \vd)$ is needed, and `C' indicates that $\vq$ is complex, requiring the most extensive solution $\texttt{LLM}(\vq, \vd, \vc)$\footnote{We consider three levels of query complexity, and leave the exploration of more fine-grained complexities as future work.}.

\paragraph{Training Strategy}
The remaining step is to train the smaller Language Model for $\texttt{Classifier}$, to accurately predict its complexity $o$ in response to the given query $\vq$. Yet, there is no annotated dataset available for query-complexity pairs. Hence, we propose to automatically construct the training dataset with two particular strategies. 

To be specific, we first aim at labeling the query complexity based on the results from three different retrieval-augmented LLM strategies, in order to determine the label by its needs. For example, if the simplest non-retrieval-based approach correctly generates the answer, the label for its corresponding query is assigned `A'. Also, to break the tie between different models in providing the label to the query, we provide a higher priority to a simpler model. In other words, if both single-step and multi-step approaches produce the same correct answer while the non-retrieval-based approach fails, we assign label `B' to its corresponding query. 

However, this labeling strategy has a limitation in that not all the queries are assigned labels, since the three retrieval-augmented approaches may all fail to generate the correct answer. On the other hand, the benchmark datasets may already have meaningful inductive biases about the most appropriate retrieval-augmented LLM strategies for their queries, considering the ways they are created (e.g., QA datasets that require sequential reasoning usually necessitate a multi-step approach; while queries of those with labeled single documents can be ideally answerable with the single-step approach). Therefore, for those queries that remain unlabeled after the first labeling step, we assign `B' to queries in single-hop datasets and `C' to queries in multi-hop datasets. Finally, we train $\texttt{Classifier}$ with these automatically-collected query-complexity pairs\footnote{As we automatically assign classifier labels, there might be errors in labeling and might be more advanced strategies to automatically assign labels, which we leave as future work.}, by using a cross-entropy loss. Then, at inference, we can determine the complexity of the query, which is one of \{`A', `B', `C'\}, by forwarding it to $\texttt{Classifier}$: $o = \texttt{Classifier}(\vq)$.

\section{Experimental Setups}
In this section, we explain datasets, models, metrics, and implementation details. We provide additional details in Appendix~\ref{appendix:setup}.

\subsection{Datasets}
In order to simulate a realistic scenario, where different queries have varying complexities, we use both the single-hop and multi-hop QA datasets simultaneously, in the unified experimental setting.

\begin{table*}[t!]
\caption{Averaged results on a collection of benchmark datasets for open-domain question answering including the single-hop and multi-hop queries, with different LLMs. Self-RAG$^*$ is trained with a different base LLM, namely LLaMA2~\cite{LLaMA2}; therefore, we compare the results of FLAN-T5-XL (3B) with the results from Self-RAG with LLaMA2 (7B) and the results of others with the results from Self-RAG with LLaMA2 (13B).  We emphasize our results in bold, for easy comparisons.}
\vspace{-0.1in}
\label{tab:main}
\small
\centering
\resizebox{\textwidth}{!}{
\renewcommand{\arraystretch}{1.0}
\begin{tabular}{llccccccccccccccc}
\toprule

& & \multicolumn{5}{c}{\bf FLAN-T5-XL (3B)} & \multicolumn{5}{c}{\bf FLAN-T5-XXL (11B)} & \multicolumn{5}{c}{\bf GPT-3.5 (Turbo)} \\
\cmidrule(l{2pt}r{2pt}){3-7} \cmidrule(l{2pt}r{2pt}){8-12} \cmidrule(l{2pt}r{2pt}){13-17}

\textbf{Types} & \textbf{Methods} & EM & F1 & Acc & Step & Time & EM & F1 & Acc & Step & Time & EM & F1 & Acc & Step & Time \\
\midrule

\multirowcell{2}[-0.0ex][l]{\textbf{Simple}} 

& \textbf{No Retrieval} & 14.87 & 21.12 & 15.97 & 0.00 & 0.11 & 17.83 & 25.14  & 19.33 & 0.00 & 0.08 & 35.77 & 48.56  & 44.27 & 0.00 & 0.71 \\

& \textbf{Single-step Approach} & 34.83 & 44.31 & 38.87 & 1.00 & 1.00 & 37.87 & 47.63  & 41.90 & 1.00 & 1.00 & 34.73 & 46.99  & 45.27 & 1.00 & 1.00 \\

\noalign{\vskip 0.25ex}\cdashline{1-17}\noalign{\vskip 0.75ex}

\multirowcell{3}[-0.0ex][l]{\textbf{Adaptive}} 

& \textbf{Adaptive Retrieval} & 23.87 & 32.24  & 26.73 & 0.50 & 0.56 & 26.93 & 35.67  & 29.73 & 0.50 & 0.54 & 35.90 & 48.20  & 45.30 & 0.50 & 0.86 \\

& \textbf{Self-RAG$^*$} & 9.90 & 20.79  & 31.57 & 0.72 & 0.43 & 10.87 & 22.98  & 34.13 & 0.74 & 0.23 & 10.87 & 22.98  & 34.13 & 0.74 & 1.50 \\

& \textbf{Adaptive-RAG (Ours)} & \textbf{37.17} & \textbf{46.94}  & \textbf{42.10} & \textbf{2.17} & \textbf{3.60} & \textbf{38.90} & \textbf{48.62} & \textbf{43.77} & \textbf{1.35} & \textbf{2.00} & \textbf{37.97} & \textbf{50.91}  & \textbf{48.97} & \textbf{1.03} & \textbf{1.46} \\

\noalign{\vskip 0.25ex}\cdashline{1-17}\noalign{\vskip 0.75ex}

\textbf{Complex}

& \textbf{Multi-step Approach} & 39.00 & 48.85  & 43.70 & 4.69 & 8.81 & 40.13 & 50.09  & 45.20 & 2.13 & 3.80 & 38.13 & 50.87  & 49.70 & 2.81 & 3.33 \\

\noalign{\vskip 0.25ex}\cdashline{1-17}\noalign{\vskip 0.75ex}

\textbf{Oracle}
& \textbf{Adaptive-RAG w/ Oracle} & 45.00 & 56.28 & 49.90 & 1.28 & 2.11 & 47.17 & 58.60 & 52.20 & 0.84 & 1.10 & 47.70 & 62.80 & 58.57 & 0.50 & 1.03 \\

\bottomrule

\end{tabular}
}
\vspace{-0.1in}
\end{table*}

\paragraph{Single-hop QA}
For simpler queries, we use three benchmark single-hop QA datasets, which consist of queries and their associated documents containing answers, namely \textbf{1) SQuAD v1.1}~\cite{squad}, \textbf{2) Natural Questions}~\cite{nq}, and \textbf{3) TriviaQA}~\cite{tqa}.

\paragraph{Multi-hop QA}
To consider more complex query scenarios, we use three benchmark multi-hop QA datasets, which require sequential reasoning over multiple documents, namely \textbf{1) MuSiQue}~\cite{musique}, \textbf{2) HotpotQA}~\cite{hotpotqa}, and \textbf{3) 2WikiMultiHopQA}~\cite{2hop}.

\begin{table*}[t!]
\caption{Results on each of a collection of datasets with FLAN-T5-XL (3B) as the LLM. We emphasize our results in bold.}
\vspace{-0.1in}
\label{tab:main:xl}
\small
\centering
\resizebox{\textwidth}{!}{
\renewcommand{\arraystretch}{0.99}
\begin{tabular}{lllccccccccccccccc}
\toprule

& & & \multicolumn{5}{c}{\bf SQuAD} & \multicolumn{5}{c}{\bf Natural Questions} & \multicolumn{5}{c}{\bf TriviaQA} \\
\cmidrule(l{2pt}r{2pt}){4-8} \cmidrule(l{2pt}r{2pt}){9-13} \cmidrule(l{2pt}r{2pt}){14-18}
\textbf{Data} & \textbf{Types} &\textbf{Methods} & EM & F1 & Acc & Step & Time & EM & F1 & Acc & Step & Time & EM & F1 & Acc & Step & Time \\

\midrule

\multirowcell{7}[-0.0ex][l]{\textbf{Single-step}} 

& \multirowcell{2}[-0.0ex][l]{\textbf{Simple}} 

& \textbf{No Retrieval} & 3.60 & 10.50 & 5.00 & 0.00 & 0.11 & 14.20 & 19.00 & 15.60 & 0.00 & 0.13 & 25.00 & 31.80 & 27.00 & 0.00 & 0.13 \\

& & \textbf{Single-step Approach} & 27.80 & 39.30 & 34.00 & 1.00 & 1.00 & 37.80 & 47.30 & 44.60 & 1.00 & 1.00 & 53.60 & 62.40 & 60.20 & 1.00 & 1.00 \\

\noalign{\vskip 0.25ex}\cdashline{2-18}\noalign{\vskip 0.75ex}

& \multirowcell{3}[-0.0ex][l]{\textbf{Adaptive}} 

& \textbf{Adaptive Retrieval} & 13.40 & 23.10 & 17.60 & 0.50 & 0.55 & 28.20 & 36.00  & 33.00 & 0.50 & 0.56 & 38.40 & 46.90 & 42.60 & 0.50 & 0.56 \\

& & \textbf{Self-RAG$^*$} & 2.20 & 11.20 & 18.40 & 0.63 & 0.50 & 31.40 & 39.00 & 33.60 & 0.63 & 0.17 & 12.80 & 29.30 & 57.00 & 0.68 & 0.45 \\

& & \textbf{Adaptive-RAG (Ours)} & \textbf{26.80} & \textbf{38.30} & \textbf{33.00} & \textbf{1.37} & \textbf{2.02} & \textbf{37.80} & \textbf{47.30}  & \textbf{44.60} & \textbf{1.00} & \textbf{1.00} & \textbf{52.20} & \textbf{60.70} & \textbf{58.20} & \textbf{1.23} & \textbf{1.54} \\

\noalign{\vskip 0.25ex}\cdashline{2-18}\noalign{\vskip 0.75ex}

& \textbf{Complex}

& \textbf{Multi-step Approach} & 24.40 & 35.60 & 29.60 & 4.52 & 9.03 & 38.60 & 47.80 & 44.20 & 5.04 & 10.18 & 53.80 & 62.40 & 60.20 & 5.28 & 9.22 \\

\noalign{\vskip 0.25ex}\cdashline{2-18}\noalign{\vskip 0.75ex}

& \textbf{Oracle}
& \textbf{Adaptive-RAG w/ Oracle} & 32.00 & 45.60 & 38.20 & 1.24 & 1.60 & 47.40 & 57.10 & 53.60 & 1.10 & 1.55 & 61.60 & 70.20 & 66.40 & 0.79 & 1.10 \\

\midrule
\midrule

& & & \multicolumn{5}{c}{\bf MuSiQue} & \multicolumn{5}{c}{\bf HotpotQA} & \multicolumn{5}{c}{\bf 2WikiMultiHopQA} \\
\cmidrule(l{2pt}r{2pt}){4-8} \cmidrule(l{2pt}r{2pt}){9-13} \cmidrule(l{2pt}r{2pt}){14-18}
\textbf{Data} & \textbf{Types} &\textbf{Methods} & EM & F1 & Acc & Step & Time & EM & F1 & Acc & Step & Time & EM & F1 & Acc & Step & Time \\

\midrule

\multirowcell{7}[-0.0ex][l]{\textbf{Multi-step}} 

& \multirowcell{2}[-0.0ex][l]{\textbf{Simple}} 

& \textbf{No Retrieval} & 2.40 & 10.70 & 3.20 & 0.00 & 0.11 & 16.60 & 22.71 & 17.20 & 0.00 & 0.11 & 27.40 & 32.04 & 27.80 & 0.00 & 0.10 \\

& & \textbf{Single-step Approach} & 13.80 & 22.80 & 15.20 & 1.00 & 1.00 & 34.40 & 46.15 & 36.40 & 1.00 & 1.00 & 41.60 & 47.90 & 42.80 & 1.00 & 1.00 \\

\noalign{\vskip 0.25ex}\cdashline{2-18}\noalign{\vskip 0.75ex}

& \multirowcell{3}[-0.0ex][l]{\textbf{Adaptive}} 

& \textbf{Adaptive Retrieval} & 6.40 & 15.80 & 8.00 & 0.50 & 0.55 & 23.60 & 32.22 & 25.00 & 0.50 & 0.55 & 33.20 & 39.44 & 34.20 & 0.50 & 0.55 \\

& & \textbf{Self-RAG$^*$} & 1.60 & 8.10 & 12.00 & 0.73 & 0.51 & 6.80 & 17.53 & 29.60 & 0.73 & 0.45 & 4.60 & 19.59 & 38.80 & 0.93 & 0.49 \\

& & \textbf{Adaptive-RAG (Ours)} & \textbf{23.60} & \textbf{31.80}  & \textbf{26.00} & \textbf{3.22} & \textbf{6.61} & \textbf{42.00} & \textbf{53.82} & \textbf{44.40} & \textbf{3.55} & \textbf{5.99} & \textbf{40.60} & \textbf{49.75} & \textbf{46.40} & \textbf{2.63} & \textbf{4.68} \\

\noalign{\vskip 0.25ex}\cdashline{2-18}\noalign{\vskip 0.75ex}

& \textbf{Complex}

& \textbf{Multi-step Approach} & 23.00 & 31.90 & 25.80 & 3.60 & 7.58 & 44.60 & 56.54 & 47.00 & 5.53 & 9.38 & 49.60 & 58.85 & 55.40 & 4.17 & 7.37 \\

\noalign{\vskip 0.25ex}\cdashline{2-18}\noalign{\vskip 0.75ex}

& \textbf{Oracle}
& \textbf{Adaptive-RAG w/ Oracle} & 24.80 & 38.50 & 27.00 & 1.98 & 3.99 & 51.20 & 64.00 & 54.80 & 1.59 & 2.77 & 53.00 & 62.30 & 59.40 & 1.01 & 1.69 \\

\bottomrule

\end{tabular}
}
\vspace{-0.11in}
\end{table*}
\begin{figure*}[t!]
    \begin{minipage}{0.33\linewidth}
        \centering
        \includegraphics[width=0.95\columnwidth]{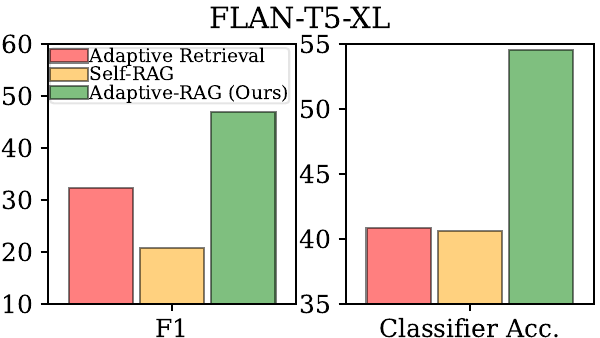}
    \end{minipage}
    \begin{minipage}{0.33\linewidth}
        \centering
        \includegraphics[width=0.95\columnwidth]{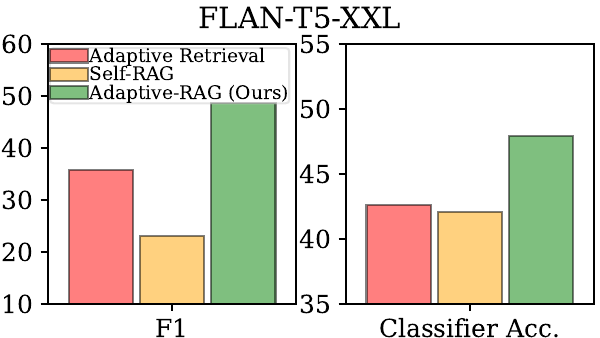}
    
    \end{minipage}
    \begin{minipage}{0.33\linewidth}
        \centering
        \includegraphics[width=0.95\columnwidth]{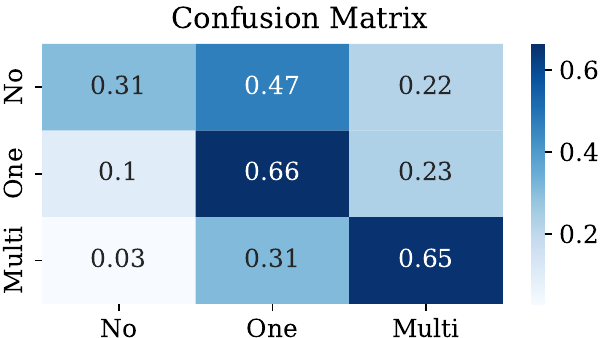}
        \label{fig:confusion_matrix}
    
    \end{minipage}
    \vspace{-0.125in}
    \caption{Performance on QA and query-complexity assessment of different adaptive approaches for retrieval-augmented LLMs with FLAN-T5 XL (Left) and XXL (Center). For labeling the complexity of queries, we use the silver data annotated from the prediction outcomes of models (described in Section~\ref{method:ours}). We also provide the confusion matrix across three labels (Right).}
    \label{fig:classifier}
    \vspace{-0.15in}
\end{figure*}

\subsection{Models}
We compare our Adaptive-RAG against relevant models, including three retrieval-augmented LLM strategies (in Section~\ref{method:pre}) and the adaptive retrieval approaches~\cite{adaptiveRetrieval, self-rag}, which can be grouped into one of three categories: Simple, Adaptive, and Complex. Specifically, Simple approaches include the \textbf{1) No Retrieval} and \textbf{2) Single-step Approach}-based methods. Adaptive approaches include the \textbf{3) Adaptive Retrieval}~\cite{adaptiveRetrieval}, \textbf{4) Self-RAG}~\cite{self-rag}, and our \textbf{5) Adaptive-RAG}, which can adaptively perform retrieval based on the question complexity. For the \textbf{6) Multi-step Approach}, we use the most sophisticated state-of-the-art method~\cite{ircot}, iteratively accessing both the retriever and LLM with Chain-of-Thought reasoning~\cite{cot}, for every query. Note that models across different categories are not directly comparable. Yet, in the ideal setting, Adaptive approaches should be more effective than those in the Simple category while simultaneously being more efficient than the Complex one. Therefore, we also report the performance in an ideal scenario, \textbf{7) Adaptive-RAG w/ Oracle}, using the oracle classifier with our Adaptive-RAG.

\subsection{Evaluation Metrics}
When it comes to evaluating adaptive models, it is essential to simultaneously consider both the task performance and efficiency along with their trade-offs. Thus, we report the results with five metrics, where three of them measure the effectiveness and the other two measure the efficiency. In particular, for effectiveness, we use F1, EM, and Accuracy (Acc), following the standard evaluation protocol~\cite{adaptiveRetrieval, KALMV, self-rag}, where F1 measures the number of overlapping words between the predicted answer and the ground truth, EM measures whether they are the same, and Acc measures whether the predicted answer contains the ground-truth answer.
For efficiency, we measure the number of retrieval-and-generate steps and the average time for answering each query relative to the one-step approach.

\subsection{Implementation Details}
For a fair comparison and following~\citet{adaptiveRetrieval} and~\citet{ircot}, we use the same retriever, a term-based sparse retrieval model known as BM25~\cite{bm25}, across all different models. For the external document corpus, we use different sources depending on the dataset type: the Wikipedia corpus preprocessed by~\citet{DPR} for single-hop datasets, and the preprocessed corpus by~\citet{ircot} for multi-hop datasets. Regarding the LLMs that are used to generate answers, we use the FLAN-T5 series models~\cite{flan} of XL with 3B parameters and XXL with 11B parameters, and the GPT-3.5 model (gpt-3.5-turbo-instruct). For the retrieval-augmented LLM design, we follow the implementation details from~\citet{ircot}, which include input prompts, instructions, and the number of test samples for evaluation (e.g., 500 samples per dataset). In our Adaptive-RAG, for the query-complexity classifier, we use and train the T5-Large model~\cite{t5}. Specifically, the classifier is trained using the epoch that shows the best performance until 100 training iterations from the validation set, with the learning rate of 3e-5 and the AdamW~\cite{adamw} as an optimizer. Regarding its training data, we sample and annotate 400 queries from 6 datasets based on its inductive bias (single-hop for one-step approach and multi-hop for multi-step). In addition, we use predicted outcomes of three different strategies over 400 queries sampled from each dataset. Note that those queries used for classifier training do not overlap with the testing queries for QA.

\section{Experimental Results and Analyses}
In this section, we show the overall experimental results and offer in-depth analyses of our method.

\paragraph{Main Results}
First of all, Table~\ref{tab:main} shows our main results averaged over all considered datasets, which corroborate our hypothesis that simple retrieval-augmented strategies are less effective than the complex strategy, while the complex one is significantly more expensive than the simple ones. In addition, we report the more granular results with FLAN-T5-XL on each of the single-hop and multi-hop datasets in Table~\ref{tab:main:xl} (and more with different LLMs in Table~\ref{tab:main:xxl} and Table~\ref{tab:main:gpt} of Appendix), which are consistent with the results observed in Table~\ref{tab:main}.

However, in a real-world scenario, not all users ask queries with the same level of complexity, which emphasizes the importance of the need for adaptive strategies. Note that among the adaptive strategies, our Adaptive-RAG shows remarkable effectiveness over the competitors (Table~\ref{tab:main}). This indicates that merely focusing on the decision of whether to retrieve or not is suboptimal. Also, as shown in Table~\ref{tab:main:xl}, such simple adaptive strategies are particularly inadequate for handling complex queries in multi-hop datasets, which require aggregated information and reasoning over multiple documents. Meanwhile, our approach can consider a more fine-grained query handling strategy by further incorporating an iterative module for complex queries. 
Furthermore, in a realistic setting, we should take into account not only effectiveness but also efficiency. As shown in Table~\ref{tab:main}, compared to the complex multi-step strategy, our proposed adaptive strategy is significantly more efficient across all model sizes. This is meaningful in this era of LLMs, where the cost of accessing them is a critical factor for practical applications and scalability.
Finally, to see the upper bound of our Adaptive-RAG, we report its performances with the oracle classifier where the classification performance is perfect. As shown in Table~\ref{tab:main} and Table~\ref{tab:main:xl}, we observe that it achieves the best performance while being much more efficient than our Adaptive-RAG without the oracle classifier. These results support the validity and significance of our proposal for adapting retrieval-augmented LLM strategies based on query complexity, and further suggest the direction to develop more improved classifiers to achieve optimal performance.

\paragraph{Classifier Performance}
To understand how the proposed classifier works, we analyze its performance across different complexity labels. As Figure~\ref{fig:classifier} (Left and Center) shows, the classification accuracy of our Adaptive-RAG is better than those of the other adaptive retrieval baselines, which leads to overall QA performance improvements. In other words, this result indicates that our Adaptive-RAG is capable of more accurately classifying the complexity levels with various granularities, which include not performing retrieval, performing retrieval only once, and performing retrieval multiple times. In addition to the true positive performance of our classifier averaged over all those three labels in Figure~\ref{fig:classifier} (Left and Center), we further report its confusion matrix in Figure~\ref{fig:classifier} (Right). We note that the confusion matrix reveals some notable trends: `C (Multi)' is sometimes misclassified as `B (One)' (about 31\%) and `B (One)' as `C (Multi)' (about 23\%); `A (No)' is misclassified often as `B (One)' (about 47\%) and less frequently as `C (Multi)' (about 22\%). While the overall results in Figure~\ref{fig:classifier} show that our classifier effectively categorizes the three labels, further refining it based on such misclassification would be a meaningful direction for future work.

\paragraph{Analyses on Efficiency for Classifier}
While Table~\ref{tab:main} shows the relative elapsed time for each of the three different RAG strategies, we further provide the exact elapsed time per query for our Adaptive-RAG and the distribution for predicted labels from our query-complexity classifier in Table~\ref{tab:classifier_efficiency}. Similar to the results of the elapsed time in Table~\ref{tab:main} (relative time), Table~\ref{tab:classifier_efficiency} (exact time) shows that efficiency can be substantially improved by identifying simple or straightforward queries.

\begin{table}[t!]
\caption{The exact elapsed time per query and the percentage of the predicted labels from the classifier over all samples.}
\vspace{-0.1in}
\label{tab:classifier_efficiency}
\small
\centering
\resizebox{0.475\textwidth}{!}{
\renewcommand{\arraystretch}{1.2}
\renewcommand{\tabcolsep}{2.5mm}
\begin{tabular}{lcc}
\toprule

\textbf{Labels} & \textbf{Time/Query (Sec.)} & \textbf{Percentage (\%)} \\

\midrule
\midrule
\multirowcell{1}[-0.0ex][l]{\textbf{No (A)}} 
& 0.35 & 8.60 \\

\multirowcell{1}[-0.0ex][l]{\textbf{One (B)}} 
& 3.08 & 53.33 \\

\multirowcell{1}[-0.0ex][l]{\textbf{Multi (C)}} 
& 27.18 & 38.07 \\

\bottomrule

\end{tabular}
}
\vspace{-0.125in}
\end{table}

\paragraph{Analyses on Training Data for Classifier}
We have shown that the classifier plays an important role in adaptive retrieval. Here, we further analyze the different strategies for training the classifier by ablating our full training strategy, which includes two approaches: generating silver data from predicted outcomes of models and utilizing inductive bias in datasets (see Section~\ref{method:ours}). As Table~\ref{tab:training_data} shows, compared to the training strategy relying solely on the data derived from inductive bias, ours is significantly more efficient. This efficiency is partly because ours also takes into account the case that does not consider any documents at all, as also implied by the classification accuracy; meanwhile, queries in the existing datasets do not capture the information on whether the retrieval is required or not. On the other hand, in the case of only using the silver data annotated from the correct predictions, while its overall classification accuracy is high, the overall QA performance implies that relying on the silver data may not be optimal. This may be because this silver data does not cover complexity labels over incorrectly predicted queries, which leads to lower generalization effect on queries relevant to them. Meanwhile, by also incorporating complexity labels from dataset bias (single-hop vs multi-hop), the classifier becomes more accurate in predicting multi-hop queries, leading to the better performance. It is worth noting that our automatic labeling strategies are two particular instantiations for training the classifier, and that there could be other instantiations, which we leave as future work.

\begin{table}[t!]
\caption{Results on QA and complexity classification with varying the data annotation strategies for training the classifier.}
\vspace{-0.1in}
\label{tab:training_data}
\small
\centering
\resizebox{0.475\textwidth}{!}{
\renewcommand{\arraystretch}{0.975}
\renewcommand{\tabcolsep}{1.0mm}
\begin{tabular}{lcccccc}
\toprule

& \multicolumn{2}{c}{\bf QA} & \multicolumn{4}{c}{\bf Classifier (Accuracy)} \\
\cmidrule(l{2pt}r{2pt}){2-3} \cmidrule(l{2pt}r{2pt}){4-7} 
\textbf{Training Strategies}  & \textbf{F1} & \textbf{Step} & \textbf{All} & \textbf{No} & \textbf{One} & \textbf{Multi} \\

\midrule
\midrule

\multirowcell{1}[-0.0ex][l]{\textbf{Adaptive-RAG (Ours)}} 

& \textbf{46.94}  &  \textbf{1084} & \textbf{54.52} & \textbf{30.52} & \textbf{66.28} &  \textbf{65.45} \\

\noalign{\vskip 0.25ex}\cdashline{1-7}\noalign{\vskip 0.75ex}

\multirowcell{1}[-0.0ex][l]{\textbf{w/o Binary}} 

&  43.43 &  640 & 60.30 & 62.19 & 65.70 & 39.55 \\

\multirowcell{1}[-0.0ex][l]{\textbf{w/o Silver}}

&  48.79 & 1464  & 40.00 & 0.00 & 53.98 & 75.91 \\

\bottomrule

\end{tabular}
}
\vspace{-0.125in}
\end{table}

\begin{table*}[t!]
\caption{Case study with GPT-3.5 (Turbo), where we represent the factual error in red and the accurate information in blue.}
\vspace{-0.1in}
\label{tab:case}
\small
\centering
\resizebox{\textwidth}{!}{
\renewcommand{\arraystretch}{1.3}
\begin{tabular}{cccc}
\toprule

\multicolumn{1}{p{.12\textwidth}}{\textbf{Dataset}} & \multicolumn{1}{p{.178\textwidth}}{\textbf{Question}} & \multicolumn{1}{p{.4\textwidth}}{\textbf{Adaptive Retrieval}} & \multicolumn{1}{p{.4\textwidth}}{\textbf{Adaptive-RAG (Ours)}} \\

\midrule
\multicolumn{1}{p{.12\textwidth}}{\textbf{NQ} \quad \; (Single-hop)}&
\multicolumn{1}{p{.178\textwidth}}{Which famous corporate logo changed to a flat colour/color sans serif font in its first major change since 1999?} &
\multicolumn{1}{p{.4\textwidth}}{\textbf{Query Type: } B (Single-step Approach)\newline \textbf{Reasoning: }The famous corporate logo that changed to a flat color sans-serif font in its first major change since 1999 is \textcolor{red}{Microsoft}. Microsoft changed its logo in 2012, replacing the italicized, multi-colored font with a simple, flat, grey font. So the answer is: \textcolor{red}{Microsoft}.} &
\multicolumn{1}{p{.4\textwidth}}{\textbf{Query Type: } A (Non Retrieval)\newline \textbf{Reasoning: }The famous corporate logo that changed to a flat color sans serif font in its first major change since 1999 is \textcolor{blue}{Google}. Google changed its logo in 2015. So the answer is: \textcolor{blue}{Google}.} \\ \midrule \midrule

\multicolumn{1}{p{.12\textwidth}}{\textbf{MuSiQue} (Multi-hop)}&
\multicolumn{1}{p{.178\textwidth}}{Who is the child of the Italian navigator who explored the eastern coast of the continent César Gaytan was born in for the English?} &
\multicolumn{1}{p{.4\textwidth}}{\textbf{Query Type: } A (Non Retrieval)\newline \textbf{Reasoning: }César Gaytan was born in the continent of \textcolor{blue}{North America}. The Italian navigator who explored the eastern coast of North America for the English was \textcolor{blue}{Giovanni Caboto}, also known as John Cabot. So the answer is: \textcolor{red}{Giovanni Caboto/John Cabot}.} &
\multicolumn{1}{p{.4\textwidth}}{\textbf{Query Type: } C (Multi-step Approach)\newline \textbf{Reasoning: }Césarr Gaytan was born in \textcolor{blue}{Guadalajara, Jalisco, Mexico}. The Italian navigator who explored the eastern coast of the continent for the English is \textcolor{blue}{John Cabot}. John Cabot's son is \textcolor{blue}{Sebastian Cabot}. So the answer is: \textcolor{blue}{Sebastian Cabot.}} \\

\bottomrule

\end{tabular}
}
\vspace{-0.125in}
\end{table*}

\paragraph{Analyses on Classifier Size}
To investigate the sensitivity of our classifier according to its varying sizes, we conducted further experiments. As shown in Table~\ref{tab:classifier_size}, we observe no significant performance differences among classifiers of various sizes, even with reduced complexity and fewer parameters in smaller classifiers. This indicates that our proposed classifier can contribute to resource-efficient settings in real-use cases with smaller sizes without compromising the performance.

\paragraph{Case Study}
We conduct a case study to qualitatively compare our Adaptive-RAG against Adaptive Retrieval. Table~\ref{tab:case} shows the classified complexity and the query handling patterns for both simple and complex questions. 
First, for the simple single-hop question, our Adaptive-RAG identifies that it is answerable by only using the LLM's parametric knowledge about `Google'. By contrast, Adaptive Retrieval fetches additional documents, leading to longer processing times and occasionally producing incorrect responses due to the inclusion of partially irrelevant information about `Microsoft'.
Meanwhile, faced with a complex question, Adaptive-RAG seeks out relevant information, including details like `a son of John Cabot', which may not have been stored in LLMs, while Adaptive Retrieval fails to request such information from external sources, resulting in inaccurate answers.

\begin{table}[t!]
\caption{Results with varying model sizes for classifiers.}
\vspace{-0.1in}
\label{tab:classifier_size}
\small
\centering
\resizebox{0.475\textwidth}{!}{
\renewcommand{\arraystretch}{1.0}
\renewcommand{\tabcolsep}{1.0mm}
\begin{tabular}{lcccccc}
\toprule

& \multicolumn{2}{c}{\bf QA} & \multicolumn{4}{c}{\bf Classifier (Accuracy)} \\
\cmidrule(l{2pt}r{2pt}){2-3} \cmidrule(l{2pt}r{2pt}){4-7} 
\textbf{Sizes}  & \textbf{F1} & \textbf{Step} & \textbf{All} & \textbf{No} & \textbf{One} & \textbf{Multi} \\

\midrule
\midrule

\multirowcell{1}[-0.0ex][l]{\textbf{Small (60M)}}

&  45.83 & 964 & 53.48  & 26.65 & 70.62 & 53.18  \\

\multirowcell{1}[-0.0ex][l]{\textbf{Base (223M)}}

& 45.97  &  983 & 53.41 & 26.42 & 69.46 & 56.82 \\

\multirowcell{1}[-0.0ex][l]{\textbf{Large (770M)}}

&  \textbf{46.94} & \textbf{1084}  & \textbf{54.52} & 30.52 & 66.28 & 65.45 \\

\bottomrule

\end{tabular}
}
\vspace{-0.1in}
\end{table}

\section{Conclusion}
In this work, we proposed the Adaptive Retrieval-Augmented Generation framework, referred to as Adaptive-RAG, to handle queries of various complexities. Specifically, Adaptive-RAG is designed to dynamically adjust its query handling strategies in the unified retrieval-augmented LLM based on the complexity of queries that they encounter, which spans across a spectrum of the non-retrieval-based approach for the most straightforward queries, to the single-step approach for the queries of moderate complexity, and finally to the multi-step approach for the complex queries. The core step of our Adaptive-RAG lies in determining the complexity of the given query, which is instrumental in selecting the most suitable strategy for its answer. To operationalize this process, we trained a smaller Language Model with query-complexity pairs, which are automatically annotated from the predicted outcomes and the inductive biases in datasets. We validated our Adaptive-RAG on a collection of open-domain QA datasets, covering the multiple query complexities including both the single- and multi-hop questions. The results demonstrate that our Adaptive-RAG enhances the overall accuracy and efficiency of QA systems, allocating more resources to handle complex queries while efficiently handling simpler queries, compared to the existing one-size-fits-all approaches that tend to be either minimalist or maximalist over varying query complexities.

\section*{Limitations}
While our Adaptive-RAG shows clear advantages in effectiveness and efficiency by determining the query complexity and then leveraging the most suitable approach for tackling it, it is important to recognize that there still exist potential avenues for improving the classifier from the perspectives of its training datasets and architecture. Specifically, as there are no available datasets for training the query-complexity classifier, we automatically create new data based on the model prediction outcomes and the inductive dataset biases. However, our labeling process is one specific instantiation of labeling the query complexity, and it may have the potential to label queries incorrectly despite its effectiveness. Therefore, future work may create new datasets that are annotated with a diverse range of query complexities, in addition to the labels of question-answer pairs. Also, as the performance gap between the ideal classifier in Table~\ref{tab:main} and the current classifier in Figure~\ref{fig:classifier} indicates, there is still room to improve the effectiveness of the classifier. In other words, our classifier design based on the smaller LM is the initial, simplest instantiation for classifying the query complexity, and based upon it, future work may improve the classifier architecture and its performance, which will positively contribute to the overall QA performance.

\section*{Ethics Statement}
The experimental results on Adaptive-RAG validate its applicability in realistic scenarios, where a wide range of diverse user queries exist. Nonetheless, given the potential diversity of real-world user inputs, it is crucial to also consider scenarios where these inputs might be offensive or harmful. We should be aware that such inputs could lead to the retrieval of offensive documents and the generation of inappropriate responses by the retrieval-augmented LLMs. To address this challenge, developing methods to detect and manage offensive or inappropriate content in both user inputs and retrieved documents within the retrieval-augmented framework is essential. We believe that this is a critical area for future work.

\section*{Acknowledgements}
This work was supported by Institute for Information and communications Technology Promotion (IITP) grant funded by the Korea government (No. 2018-0-00582, Prediction and augmentation of the credibility distribution via linguistic analysis and automated evidence document collection), Basic Science Research Program through the National Research Foundation of Korea (NRF) funded by the Ministry of Education (RS-2023-00275747), and the Artificial intelligence industrial convergence cluster development project funded by the Ministry of Science and ICT (MSIT, Korea) \& Gwangju Metropolitan City.

\bibliography{anthology,custom}

\begin{thebibliography}{45}
\expandafter\ifx\csname natexlab\endcsname\relax\def\natexlab#1{#1}\fi

\bibitem[{Anil et~al.(2023)Anil, Dai, Firat, Johnson, Lepikhin, Passos, Shakeri, Taropa, Bailey, Chen, Chu, Clark, Shafey, Huang, Meier{-}Hellstern, Mishra, Moreira, Omernick, Robinson, Ruder, Tay, Xiao, Xu, Zhang, {\'{A}}brego, Ahn, Austin, Barham, Botha, Bradbury, Brahma, Brooks, Catasta, Cheng, Cherry, Choquette{-}Choo, Chowdhery, Crepy, Dave, Dehghani, Dev, Devlin, D{\'{\i}}az, Du, Dyer, Feinberg, Feng, Fienber, Freitag, Garcia, Gehrmann, Gonzalez, and et~al.}]{PaLM2}
Rohan Anil, Andrew~M. Dai, Orhan Firat, Melvin Johnson, Dmitry Lepikhin, Alexandre Passos, Siamak Shakeri, Emanuel Taropa, Paige Bailey, Zhifeng Chen, Eric Chu, Jonathan~H. Clark, Laurent~El Shafey, Yanping Huang, Kathy Meier{-}Hellstern, Gaurav Mishra, Erica Moreira, Mark Omernick, Kevin Robinson, Sebastian Ruder, Yi~Tay, Kefan Xiao, Yuanzhong Xu, Yujing Zhang, Gustavo~Hern{\'{a}}ndez {\'{A}}brego, Junwhan Ahn, Jacob Austin, Paul Barham, Jan~A. Botha, James Bradbury, Siddhartha Brahma, Kevin Brooks, Michele Catasta, Yong Cheng, Colin Cherry, Christopher~A. Choquette{-}Choo, Aakanksha Chowdhery, Cl{\'{e}}ment Crepy, Shachi Dave, Mostafa Dehghani, Sunipa Dev, Jacob Devlin, Mark D{\'{\i}}az, Nan Du, Ethan Dyer, Vladimir Feinberg, Fangxiaoyu Feng, Vlad Fienber, Markus Freitag, Xavier Garcia, Sebastian Gehrmann, Lucas Gonzalez, and et~al. 2023.
\newblock \href {https://doi.org/10.48550/ARXIV.2305.10403} {Palm 2 technical report}.
\newblock \emph{arXiv preprint arXiv:2305.10403}.

\bibitem[{Asai et~al.(2024)Asai, Wu, Wang, Sil, and Hajishirzi}]{self-rag}
Akari Asai, Zeqiu Wu, Yizhong Wang, Avirup Sil, and Hannaneh Hajishirzi. 2024.
\newblock \href {https://openreview.net/forum?id=hSyW5go0v8} {Self-{RAG}: Learning to retrieve, generate, and critique through self-reflection}.
\newblock In \emph{The Twelfth International Conference on Learning Representations}.

\bibitem[{Baek et~al.(2023)Baek, Jeong, Kang, Park, and Hwang}]{KALMV}
Jinheon Baek, Soyeong Jeong, Minki Kang, Jong Park, and Sung~Ju Hwang. 2023.
\newblock \href {https://aclanthology.org/2023.emnlp-main.107} {Knowledge-augmented language model verification}.
\newblock In \emph{Proceedings of the 2023 Conference on Empirical Methods in Natural Language Processing, {EMNLP} 2023, Singapore, December 6-10, 2023}, pages 1720--1736. Association for Computational Linguistics.

\bibitem[{Borgeaud et~al.(2022)Borgeaud, Mensch, Hoffmann, Cai, Rutherford, Millican, van~den Driessche, Lespiau, Damoc, Clark, de~Las~Casas, Guy, Menick, Ring, Hennigan, Huang, Maggiore, Jones, Cassirer, Brock, Paganini, Irving, Vinyals, Osindero, Simonyan, Rae, Elsen, and Sifre}]{retro}
Sebastian Borgeaud, Arthur Mensch, Jordan Hoffmann, Trevor Cai, Eliza Rutherford, Katie Millican, George van~den Driessche, Jean{-}Baptiste Lespiau, Bogdan Damoc, Aidan Clark, Diego de~Las~Casas, Aurelia Guy, Jacob Menick, Roman Ring, Tom Hennigan, Saffron Huang, Loren Maggiore, Chris Jones, Albin Cassirer, Andy Brock, Michela Paganini, Geoffrey Irving, Oriol Vinyals, Simon Osindero, Karen Simonyan, Jack~W. Rae, Erich Elsen, and Laurent Sifre. 2022.
\newblock \href {https://proceedings.mlr.press/v162/borgeaud22a.html} {Improving language models by retrieving from trillions of tokens}.
\newblock In \emph{International Conference on Machine Learning, {ICML} 2022, 17-23 July 2022, Baltimore, Maryland, {USA}}, volume 162 of \emph{Proceedings of Machine Learning Research}, pages 2206--2240. {PMLR}.

\bibitem[{Brown et~al.(2020)Brown, Mann, Ryder, Subbiah, Kaplan, Dhariwal, Neelakantan, Shyam, Sastry, Askell, Agarwal, Herbert{-}Voss, Krueger, Henighan, Child, Ramesh, Ziegler, Wu, Winter, Hesse, Chen, Sigler, Litwin, Gray, Chess, Clark, Berner, McCandlish, Radford, Sutskever, and Amodei}]{GPT-3}
Tom~B. Brown, Benjamin Mann, Nick Ryder, Melanie Subbiah, Jared Kaplan, Prafulla Dhariwal, Arvind Neelakantan, Pranav Shyam, Girish Sastry, Amanda Askell, Sandhini Agarwal, Ariel Herbert{-}Voss, Gretchen Krueger, Tom Henighan, Rewon Child, Aditya Ramesh, Daniel~M. Ziegler, Jeffrey Wu, Clemens Winter, Christopher Hesse, Mark Chen, Eric Sigler, Mateusz Litwin, Scott Gray, Benjamin Chess, Jack Clark, Christopher Berner, Sam McCandlish, Alec Radford, Ilya Sutskever, and Dario Amodei. 2020.
\newblock \href {https://proceedings.neurips.cc/paper/2020/hash/1457c0d6bfcb4967418bfb8ac142f64a-Abstract.html} {Language models are few-shot learners}.
\newblock In \emph{Advances in Neural Information Processing Systems 33: Annual Conference on Neural Information Processing Systems 2020, NeurIPS 2020, December 6-12, 2020, virtual}.

\bibitem[{Chen et~al.(2017)Chen, Fisch, Weston, and Bordes}]{DBLP:conf/acl/ChenFWB17}
Danqi Chen, Adam Fisch, Jason Weston, and Antoine Bordes. 2017.
\newblock \href {https://doi.org/10.18653/V1/P17-1171} {Reading wikipedia to answer open-domain questions}.
\newblock In \emph{Proceedings of the 55th Annual Meeting of the Association for Computational Linguistics, {ACL} 2017, Vancouver, Canada, July 30 - August 4, Volume 1: Long Papers}, pages 1870--1879. Association for Computational Linguistics.

\bibitem[{Cho et~al.(2023)Cho, Seo, Jeong, and Park}]{DBLP:conf/emnlp/ChoSJP23}
Sukmin Cho, Jeongyeon Seo, Soyeong Jeong, and Jong~C. Park. 2023.
\newblock \href {https://aclanthology.org/2023.findings-emnlp.207} {Improving zero-shot reader by reducing distractions from irrelevant documents in open-domain question answering}.
\newblock In \emph{Findings of the Association for Computational Linguistics: {EMNLP} 2023, Singapore, December 6-10, 2023}, pages 3145--3157. Association for Computational Linguistics.

\bibitem[{Chung et~al.(2022)Chung, Hou, Longpre, Zoph, Tay, Fedus, Li, Wang, Dehghani, Brahma, Webson, Gu, Dai, Suzgun, Chen, Chowdhery, Narang, Mishra, Yu, Zhao, Huang, Dai, Yu, Petrov, Chi, Dean, Devlin, Roberts, Zhou, Le, and Wei}]{flan}
Hyung~Won Chung, Le~Hou, Shayne Longpre, Barret Zoph, Yi~Tay, William Fedus, Eric Li, Xuezhi Wang, Mostafa Dehghani, Siddhartha Brahma, Albert Webson, Shixiang~Shane Gu, Zhuyun Dai, Mirac Suzgun, Xinyun Chen, Aakanksha Chowdhery, Sharan Narang, Gaurav Mishra, Adams Yu, Vincent~Y. Zhao, Yanping Huang, Andrew~M. Dai, Hongkun Yu, Slav Petrov, Ed~H. Chi, Jeff Dean, Jacob Devlin, Adam Roberts, Denny Zhou, Quoc~V. Le, and Jason Wei. 2022.
\newblock \href {https://doi.org/10.48550/ARXIV.2210.11416} {Scaling instruction-finetuned language models}.
\newblock \emph{arXiv preprint arXiv:2210.11416}.

\bibitem[{Ho et~al.(2020)Ho, Nguyen, Sugawara, and Aizawa}]{2hop}
Xanh Ho, Anh{-}Khoa~Duong Nguyen, Saku Sugawara, and Akiko Aizawa. 2020.
\newblock \href {https://doi.org/10.18653/V1/2020.COLING-MAIN.580} {Constructing {A} multi-hop {QA} dataset for comprehensive evaluation of reasoning steps}.
\newblock In \emph{Proceedings of the 28th International Conference on Computational Linguistics, {COLING} 2020, Barcelona, Spain (Online), December 8-13, 2020}, pages 6609--6625. International Committee on Computational Linguistics.

\bibitem[{Izacard and Grave(2021)}]{FiD}
Gautier Izacard and Edouard Grave. 2021.
\newblock \href {https://doi.org/10.18653/V1/2021.EACL-MAIN.74} {Leveraging passage retrieval with generative models for open domain question answering}.
\newblock In \emph{Proceedings of the 16th Conference of the European Chapter of the Association for Computational Linguistics: Main Volume, {EACL} 2021, Online, April 19 - 23, 2021}, pages 874--880. Association for Computational Linguistics.

\bibitem[{Izacard et~al.(2023)Izacard, Lewis, Lomeli, Hosseini, Petroni, Schick, Dwivedi{-}Yu, Joulin, Riedel, and Grave}]{atlas}
Gautier Izacard, Patrick S.~H. Lewis, Maria Lomeli, Lucas Hosseini, Fabio Petroni, Timo Schick, Jane Dwivedi{-}Yu, Armand Joulin, Sebastian Riedel, and Edouard Grave. 2023.
\newblock \href {http://jmlr.org/papers/v24/23-0037.html} {Atlas: Few-shot learning with retrieval augmented language models}.
\newblock \emph{J. Mach. Learn. Res.}, 24:251:1--251:43.

\bibitem[{Jeong et~al.(2023)Jeong, Baek, Cho, Hwang, and Park}]{DBLP:conf/emnlp/JeongBCHP23}
Soyeong Jeong, Jinheon Baek, Sukmin Cho, Sung~Ju Hwang, and Jong Park. 2023.
\newblock \href {https://aclanthology.org/2023.findings-emnlp.1033} {Test-time self-adaptive small language models for question answering}.
\newblock In \emph{Findings of the Association for Computational Linguistics: {EMNLP} 2023, Singapore, December 6-10, 2023}, pages 15459--15469. Association for Computational Linguistics.

\bibitem[{Jiang et~al.(2023)Jiang, Xu, Gao, Sun, Liu, Dwivedi{-}Yu, Yang, Callan, and Neubig}]{ActiveRetrieval}
Zhengbao Jiang, Frank~F. Xu, Luyu Gao, Zhiqing Sun, Qian Liu, Jane Dwivedi{-}Yu, Yiming Yang, Jamie Callan, and Graham Neubig. 2023.
\newblock \href {https://doi.org/10.48550/ARXIV.2305.06983} {Active retrieval augmented generation}.
\newblock In \emph{EMNLP 2023}.

\bibitem[{Joshi et~al.(2017)Joshi, Choi, Weld, and Zettlemoyer}]{tqa}
Mandar Joshi, Eunsol Choi, Daniel~S. Weld, and Luke Zettlemoyer. 2017.
\newblock \href {https://doi.org/10.18653/V1/P17-1147} {Triviaqa: {A} large scale distantly supervised challenge dataset for reading comprehension}.
\newblock In \emph{Proceedings of the 55th Annual Meeting of the Association for Computational Linguistics, {ACL} 2017, Vancouver, Canada, July 30 - August 4, Volume 1: Long Papers}, pages 1601--1611. Association for Computational Linguistics.

\bibitem[{Karpukhin et~al.(2020)Karpukhin, Oguz, Min, Lewis, Wu, Edunov, Chen, and Yih}]{DPR}
Vladimir Karpukhin, Barlas Oguz, Sewon Min, Patrick S.~H. Lewis, Ledell Wu, Sergey Edunov, Danqi Chen, and Wen{-}tau Yih. 2020.
\newblock \href {https://doi.org/10.18653/v1/2020.emnlp-main.550} {Dense passage retrieval for open-domain question answering}.
\newblock In \emph{Proceedings of the 2020 Conference on Empirical Methods in Natural Language Processing, {EMNLP} 2020, November 16-20, 2020}. Association for Computational Linguistics.

\bibitem[{Kasai et~al.(2022)Kasai, Sakaguchi, Takahashi, Bras, Asai, Yu, Radev, Smith, Choi, and Inui}]{RQA}
Jungo Kasai, Keisuke Sakaguchi, Yoichi Takahashi, Ronan~Le Bras, Akari Asai, Xinyan Yu, Dragomir~R. Radev, Noah~A. Smith, Yejin Choi, and Kentaro Inui. 2022.
\newblock \href {https://doi.org/10.48550/ARXIV.2207.13332} {Realtime {QA:} what's the answer right now?}
\newblock \emph{arXiv preprint arXiv:2207.13332}.

\bibitem[{Khattab et~al.(2022)Khattab, Santhanam, Li, Hall, Liang, Potts, and Zaharia}]{dsp}
Omar Khattab, Keshav Santhanam, Xiang~Lisa Li, David Hall, Percy Liang, Christopher Potts, and Matei Zaharia. 2022.
\newblock \href {https://doi.org/10.48550/ARXIV.2212.14024} {Demonstrate-search-predict: Composing retrieval and language models for knowledge-intensive {NLP}}.
\newblock \emph{arXiv preprint arXiv.2212.14024}, abs/2212.14024.

\bibitem[{Khot et~al.(2023)Khot, Trivedi, Finlayson, Fu, Richardson, Clark, and Sabharwal}]{decomp}
Tushar Khot, Harsh Trivedi, Matthew Finlayson, Yao Fu, Kyle Richardson, Peter Clark, and Ashish Sabharwal. 2023.
\newblock \href {https://openreview.net/pdf?id=\_nGgzQjzaRy} {Decomposed prompting: {A} modular approach for solving complex tasks}.
\newblock In \emph{The Eleventh International Conference on Learning Representations, {ICLR} 2023, Kigali, Rwanda, May 1-5, 2023}. OpenReview.net.

\bibitem[{Kwiatkowski et~al.(2019)Kwiatkowski, Palomaki, Redfield, Collins, Parikh, Alberti, Epstein, Polosukhin, Devlin, Lee, Toutanova, Jones, Kelcey, Chang, Dai, Uszkoreit, Le, and Petrov}]{nq}
Tom Kwiatkowski, Jennimaria Palomaki, Olivia Redfield, Michael Collins, Ankur Parikh, Chris Alberti, Danielle Epstein, Illia Polosukhin, Jacob Devlin, Kenton Lee, Kristina Toutanova, Llion Jones, Matthew Kelcey, Ming-Wei Chang, Andrew~M. Dai, Jakob Uszkoreit, Quoc Le, and Slav Petrov. 2019.
\newblock \href {https://doi.org/10.1162/tacl_a_00276} {Natural questions: A benchmark for question answering research}.
\newblock \emph{Transactions of the Association for Computational Linguistics}, 7:452--466.

\bibitem[{Lazaridou et~al.(2022)Lazaridou, Gribovskaya, Stokowiec, and Grigorev}]{internet-QA}
Angeliki Lazaridou, Elena Gribovskaya, Wojciech Stokowiec, and Nikolai Grigorev. 2022.
\newblock \href {https://doi.org/10.48550/ARXIV.2203.05115} {Internet-augmented language models through few-shot prompting for open-domain question answering}.
\newblock \emph{arXiv preprint arXiv:2203.05115}.

\bibitem[{Li et~al.(2020)Li, Min, Iyer, Mehdad, and Yih}]{blinke}
Belinda~Z. Li, Sewon Min, Srinivasan Iyer, Yashar Mehdad, and Wen{-}tau Yih. 2020.
\newblock \href {https://doi.org/10.18653/V1/2020.EMNLP-MAIN.522} {Efficient one-pass end-to-end entity linking for questions}.
\newblock In \emph{Proceedings of the 2020 Conference on Empirical Methods in Natural Language Processing, {EMNLP} 2020, Online, November 16-20, 2020}, pages 6433--6441. Association for Computational Linguistics.

\bibitem[{Loshchilov and Hutter(2019)}]{adamw}
Ilya Loshchilov and Frank Hutter. 2019.
\newblock \href {https://openreview.net/forum?id=Bkg6RiCqY7} {Decoupled weight decay regularization}.
\newblock In \emph{7th International Conference on Learning Representations, {ICLR} 2019, New Orleans, LA, USA, May 6-9, 2019}. OpenReview.net.

\bibitem[{Mallen et~al.(2023)Mallen, Asai, Zhong, Das, Khashabi, and Hajishirzi}]{adaptiveRetrieval}
Alex Mallen, Akari Asai, Victor Zhong, Rajarshi Das, Daniel Khashabi, and Hannaneh Hajishirzi. 2023.
\newblock \href {https://doi.org/10.18653/V1/2023.ACL-LONG.546} {When not to trust language models: Investigating effectiveness of parametric and non-parametric memories}.
\newblock In \emph{Proceedings of the 61st Annual Meeting of the Association for Computational Linguistics (Volume 1: Long Papers), {ACL} 2023, Toronto, Canada, July 9-14, 2023}, pages 9802--9822. Association for Computational Linguistics.

\bibitem[{OpenAI(2023)}]{GPT-4}
OpenAI. 2023.
\newblock \href {https://doi.org/10.48550/ARXIV.2303.08774} {{GPT-4} technical report}.
\newblock \emph{arXiv preprint arXiv:2303.08774}.

\bibitem[{Paszke et~al.(2019)Paszke, Gross, Massa, Lerer, Bradbury, Chanan, Killeen, Lin, Gimelshein, Antiga, Desmaison, K{\"{o}}pf, Yang, DeVito, Raison, Tejani, Chilamkurthy, Steiner, Fang, Bai, and Chintala}]{DBLP:conf/nips/PaszkeGMLBCKLGA19}
Adam Paszke, Sam Gross, Francisco Massa, Adam Lerer, James Bradbury, Gregory Chanan, Trevor Killeen, Zeming Lin, Natalia Gimelshein, Luca Antiga, Alban Desmaison, Andreas K{\"{o}}pf, Edward~Z. Yang, Zachary DeVito, Martin Raison, Alykhan Tejani, Sasank Chilamkurthy, Benoit Steiner, Lu~Fang, Junjie Bai, and Soumith Chintala. 2019.
\newblock \href {https://proceedings.neurips.cc/paper/2019/hash/bdbca288fee7f92f2bfa9f7012727740-Abstract.html} {Pytorch: An imperative style, high-performance deep learning library}.
\newblock In \emph{Advances in Neural Information Processing Systems 32: Annual Conference on Neural Information Processing Systems 2019}, pages 8024--8035.

\bibitem[{Pereira et~al.(2023)Pereira, do~Nascimento~Fidalgo, de~Alencar~Lotufo, and Nogueira}]{DBLP:conf/ecir/PereiraFLN23}
Jayr~Alencar Pereira, Robson do~Nascimento~Fidalgo, Roberto de~Alencar~Lotufo, and Rodrigo~Frassetto Nogueira. 2023.
\newblock \href {https://doi.org/10.1007/978-3-031-28238-6\_44} {Visconde: Multi-document {QA} with {GPT-3} and neural reranking}.
\newblock In \emph{Advances in Information Retrieval - 45th European Conference on Information Retrieval, {ECIR} 2023, Dublin, Ireland, April 2-6, 2023, Proceedings, Part {II}}, volume 13981 of \emph{Lecture Notes in Computer Science}, pages 534--543. Springer.

\bibitem[{Press et~al.(2023)Press, Zhang, Min, Schmidt, Smith, and Lewis}]{self-ask}
Ofir Press, Muru Zhang, Sewon Min, Ludwig Schmidt, Noah~A. Smith, and Mike Lewis. 2023.
\newblock \href {https://doi.org/10.48550/ARXIV.2210.03350} {Measuring and narrowing the compositionality gap in language models}.
\newblock In \emph{Findings of the Association for Computational Linguistics: EMNLP 2023}.

\bibitem[{Qi et~al.(2021)Qi, Lee, Sido, and Manning}]{DBLP:conf/emnlp/0003LSM21}
Peng Qi, Haejun Lee, Tg~Sido, and Christopher~D. Manning. 2021.
\newblock \href {https://doi.org/10.18653/V1/2021.EMNLP-MAIN.292} {Answering open-domain questions of varying reasoning steps from text}.
\newblock In \emph{Proceedings of the 2021 Conference on Empirical Methods in Natural Language Processing, {EMNLP} 2021, Virtual Event / Punta Cana, Dominican Republic, 7-11 November, 2021}, pages 3599--3614. Association for Computational Linguistics.

\bibitem[{Raffel et~al.(2020)Raffel, Shazeer, Roberts, Lee, Narang, Matena, Zhou, Li, and Liu}]{t5}
Colin Raffel, Noam Shazeer, Adam Roberts, Katherine Lee, Sharan Narang, Michael Matena, Yanqi Zhou, Wei Li, and Peter~J. Liu. 2020.
\newblock \href {http://jmlr.org/papers/v21/20-074.html} {Exploring the limits of transfer learning with a unified text-to-text transformer}.
\newblock \emph{J. Mach. Learn. Res.}, 21:140:1--140:67.

\bibitem[{Rajpurkar et~al.(2016)Rajpurkar, Zhang, Lopyrev, and Liang}]{squad}
Pranav Rajpurkar, Jian Zhang, Konstantin Lopyrev, and Percy Liang. 2016.
\newblock \href {https://doi.org/10.18653/V1/D16-1264} {Squad: 100, 000+ questions for machine comprehension of text}.
\newblock In \emph{Proceedings of the 2016 Conference on Empirical Methods in Natural Language Processing, {EMNLP} 2016, Austin, Texas, USA, November 1-4, 2016}, pages 2383--2392. The Association for Computational Linguistics.

\bibitem[{Ram et~al.(2023)Ram, Levine, Dalmedigos, Muhlgay, Shashua, Leyton-Brown, and Shoham}]{ram2023ralm}
Ori Ram, Yoav Levine, Itay Dalmedigos, Dor Muhlgay, Amnon Shashua, Kevin Leyton-Brown, and Yoav Shoham. 2023.
\newblock \href {https://arxiv.org/abs/2302.00083} {In-context retrieval-augmented language models}.
\newblock \emph{Transactions of the Association for Computational Linguistics}.

\bibitem[{Robertson et~al.(1994)Robertson, Walker, Jones, Hancock{-}Beaulieu, and Gatford}]{bm25}
Stephen~E. Robertson, Steve Walker, Susan Jones, Micheline Hancock{-}Beaulieu, and Mike Gatford. 1994.
\newblock \href {http://trec.nist.gov/pubs/trec3/papers/city.ps.gz} {Okapi at {TREC-3}}.
\newblock In \emph{Proceedings of The Third Text REtrieval Conference, {TREC} 1994, Gaithersburg, Maryland, USA, November 2-4, 1994}, volume 500-225 of \emph{{NIST} Special Publication}, pages 109--126. National Institute of Standards and Technology {(NIST)}.

\bibitem[{Shi et~al.(2023)Shi, Min, Yasunaga, Seo, James, Lewis, Zettlemoyer, and Yih}]{replug}
Weijia Shi, Sewon Min, Michihiro Yasunaga, Minjoon Seo, Rich James, Mike Lewis, Luke Zettlemoyer, and Wen{-}tau Yih. 2023.
\newblock \href {https://doi.org/10.48550/ARXIV.2301.12652} {{REPLUG:} retrieval-augmented black-box language models}.
\newblock \emph{arXiv preprint arXiv:2301.12652}.

\bibitem[{Touvron et~al.(2023)Touvron, Martin, Stone, Albert, Almahairi, Babaei, Bashlykov, Batra, Bhargava, Bhosale, Bikel, Blecher, Canton{-}Ferrer, Chen, Cucurull, Esiobu, Fernandes, Fu, Fu, Fuller, Gao, Goswami, Goyal, Hartshorn, Hosseini, Hou, Inan, Kardas, Kerkez, Khabsa, Kloumann, Korenev, Koura, Lachaux, Lavril, Lee, Liskovich, Lu, Mao, Martinet, Mihaylov, Mishra, Molybog, Nie, Poulton, Reizenstein, Rungta, Saladi, Schelten, Silva, Smith, Subramanian, Tan, Tang, Taylor, Williams, Kuan, Xu, Yan, Zarov, Zhang, Fan, Kambadur, Narang, Rodriguez, Stojnic, Edunov, and Scialom}]{LLaMA2}
Hugo Touvron, Louis Martin, Kevin Stone, Peter Albert, Amjad Almahairi, Yasmine Babaei, Nikolay Bashlykov, Soumya Batra, Prajjwal Bhargava, Shruti Bhosale, Dan Bikel, Lukas Blecher, Cristian Canton{-}Ferrer, Moya Chen, Guillem Cucurull, David Esiobu, Jude Fernandes, Jeremy Fu, Wenyin Fu, Brian Fuller, Cynthia Gao, Vedanuj Goswami, Naman Goyal, Anthony Hartshorn, Saghar Hosseini, Rui Hou, Hakan Inan, Marcin Kardas, Viktor Kerkez, Madian Khabsa, Isabel Kloumann, Artem Korenev, Punit~Singh Koura, Marie{-}Anne Lachaux, Thibaut Lavril, Jenya Lee, Diana Liskovich, Yinghai Lu, Yuning Mao, Xavier Martinet, Todor Mihaylov, Pushkar Mishra, Igor Molybog, Yixin Nie, Andrew Poulton, Jeremy Reizenstein, Rashi Rungta, Kalyan Saladi, Alan Schelten, Ruan Silva, Eric~Michael Smith, Ranjan Subramanian, Xiaoqing~Ellen Tan, Binh Tang, Ross Taylor, Adina Williams, Jian~Xiang Kuan, Puxin Xu, Zheng Yan, Iliyan Zarov, Yuchen Zhang, Angela Fan, Melanie Kambadur, Sharan Narang, Aur{\'{e}}lien Rodriguez, Robert Stojnic, Sergey Edunov,
  and Thomas Scialom. 2023.
\newblock \href {https://doi.org/10.48550/ARXIV.2307.09288} {Llama 2: Open foundation and fine-tuned chat models}.
\newblock \emph{arXiv preprint arXiv:2307.09288}.

\bibitem[{Trivedi et~al.(2022{\natexlab{a}})Trivedi, Balasubramanian, Khot, and Sabharwal}]{musique}
Harsh Trivedi, Niranjan Balasubramanian, Tushar Khot, and Ashish Sabharwal. 2022{\natexlab{a}}.
\newblock \href {https://doi.org/10.1162/TACL\_A\_00475} {Musique: Multihop questions via single-hop question composition}.
\newblock \emph{Trans. Assoc. Comput. Linguistics}, 10:539--554.

\bibitem[{Trivedi et~al.(2022{\natexlab{b}})Trivedi, Balasubramanian, Khot, and Sabharwal}]{trivedi-etal-2022-musique}
Harsh Trivedi, Niranjan Balasubramanian, Tushar Khot, and Ashish Sabharwal. 2022{\natexlab{b}}.
\newblock \href {https://doi.org/10.1162/tacl_a_00475} {♫ {M}u{S}i{Q}ue: Multihop questions via single-hop question composition}.
\newblock \emph{Transactions of the Association for Computational Linguistics}, 10:539--554.

\bibitem[{Trivedi et~al.(2023)Trivedi, Balasubramanian, Khot, and Sabharwal}]{ircot}
Harsh Trivedi, Niranjan Balasubramanian, Tushar Khot, and Ashish Sabharwal. 2023.
\newblock \href {https://doi.org/10.18653/V1/2023.ACL-LONG.557} {Interleaving retrieval with chain-of-thought reasoning for knowledge-intensive multi-step questions}.
\newblock In \emph{Proceedings of the 61st Annual Meeting of the Association for Computational Linguistics (Volume 1: Long Papers), {ACL} 2023, Toronto, Canada, July 9-14, 2023}, pages 10014--10037. Association for Computational Linguistics.

\bibitem[{Wei et~al.(2022{\natexlab{a}})Wei, Tay, Bommasani, Raffel, Zoph, Borgeaud, Yogatama, Bosma, Zhou, Metzler, Chi, Hashimoto, Vinyals, Liang, Dean, and Fedus}]{EmergentLLMs}
Jason Wei, Yi~Tay, Rishi Bommasani, Colin Raffel, Barret Zoph, Sebastian Borgeaud, Dani Yogatama, Maarten Bosma, Denny Zhou, Donald Metzler, Ed~H. Chi, Tatsunori Hashimoto, Oriol Vinyals, Percy Liang, Jeff Dean, and William Fedus. 2022{\natexlab{a}}.
\newblock \href {https://openreview.net/forum?id=yzkSU5zdwD} {Emergent abilities of large language models}.
\newblock \emph{Trans. Mach. Learn. Res.}, 2022.

\bibitem[{Wei et~al.(2022{\natexlab{b}})Wei, Wang, Schuurmans, Bosma, Ichter, Xia, Chi, Le, and Zhou}]{cot}
Jason Wei, Xuezhi Wang, Dale Schuurmans, Maarten Bosma, Brian Ichter, Fei Xia, Ed~H. Chi, Quoc~V. Le, and Denny Zhou. 2022{\natexlab{b}}.
\newblock \href {http://papers.nips.cc/paper\_files/paper/2022/hash/9d5609613524ecf4f15af0f7b31abca4-Abstract-Conference.html} {Chain-of-thought prompting elicits reasoning in large language models}.
\newblock In \emph{NeurIPS}.

\bibitem[{Wolf et~al.(2020)Wolf, Debut, Sanh, Chaumond, Delangue, Moi, Cistac, Rault, Louf, Funtowicz, Davison, Shleifer, von Platen, Ma, Jernite, Plu, Xu, Scao, Gugger, Drame, Lhoest, and Rush}]{DBLP:conf/emnlp/WolfDSCDMCRLFDS20}
Thomas Wolf, Lysandre Debut, Victor Sanh, Julien Chaumond, Clement Delangue, Anthony Moi, Pierric Cistac, Tim Rault, R{\'{e}}mi Louf, Morgan Funtowicz, Joe Davison, Sam Shleifer, Patrick von Platen, Clara Ma, Yacine Jernite, Julien Plu, Canwen Xu, Teven~Le Scao, Sylvain Gugger, Mariama Drame, Quentin Lhoest, and Alexander~M. Rush. 2020.
\newblock \href {https://doi.org/10.18653/v1/2020.emnlp-demos.6} {Transformers: State-of-the-art natural language processing}.
\newblock In \emph{Proceedings of the 2020 Conference on Empirical Methods in Natural Language Processing: System Demonstrations, {EMNLP} 2020 - Demos}, pages 38--45. Association for Computational Linguistics.

\bibitem[{Xiong et~al.(2021)Xiong, Xiong, Li, Tang, Liu, Bennett, Ahmed, and Overwijk}]{ANCE}
Lee Xiong, Chenyan Xiong, Ye~Li, Kwok{-}Fung Tang, Jialin Liu, Paul~N. Bennett, Junaid Ahmed, and Arnold Overwijk. 2021.
\newblock \href {https://openreview.net/forum?id=zeFrfgyZln} {Approximate nearest neighbor negative contrastive learning for dense text retrieval}.
\newblock In \emph{9th International Conference on Learning Representations, {ICLR} 2021, Virtual Event, Austria, May 3-7, 2021}. OpenReview.net.

\bibitem[{Yang et~al.(2019)Yang, Xie, Lin, Li, Tan, Xiong, Li, and Lin}]{BERTserini}
Wei Yang, Yuqing Xie, Aileen Lin, Xingyu Li, Luchen Tan, Kun Xiong, Ming Li, and Jimmy Lin. 2019.
\newblock \href {https://doi.org/10.18653/V1/N19-4013} {End-to-end open-domain question answering with bertserini}.
\newblock In \emph{Proceedings of the 2019 Conference of the North American Chapter of the Association for Computational Linguistics: Human Language Technologies, {NAACL-HLT} 2019, Minneapolis, MN, USA, June 2-7, 2019, Demonstrations}, pages 72--77. Association for Computational Linguistics.

\bibitem[{Yang et~al.(2018)Yang, Qi, Zhang, Bengio, Cohen, Salakhutdinov, and Manning}]{hotpotqa}
Zhilin Yang, Peng Qi, Saizheng Zhang, Yoshua Bengio, William Cohen, Ruslan Salakhutdinov, and Christopher~D. Manning. 2018.
\newblock \href {https://doi.org/10.18653/v1/D18-1259} {{H}otpot{QA}: A dataset for diverse, explainable multi-hop question answering}.
\newblock In \emph{Proceedings of the 2018 Conference on Empirical Methods in Natural Language Processing}, pages 2369--2380, Brussels, Belgium. Association for Computational Linguistics.

\bibitem[{Yao et~al.(2023)Yao, Zhao, Yu, Du, Shafran, Narasimhan, and Cao}]{react}
Shunyu Yao, Jeffrey Zhao, Dian Yu, Nan Du, Izhak Shafran, Karthik~R. Narasimhan, and Yuan Cao. 2023.
\newblock \href {https://openreview.net/pdf?id=WE\_vluYUL-X} {React: Synergizing reasoning and acting in language models}.
\newblock In \emph{The Eleventh International Conference on Learning Representations, {ICLR} 2023, Kigali, Rwanda, May 1-5, 2023}. OpenReview.net.

\bibitem[{Zhu et~al.(2021)Zhu, Lei, Wang, Zheng, Poria, and Chua}]{ODQA/survey}
Fengbin Zhu, Wenqiang Lei, Chao Wang, Jianming Zheng, Soujanya Poria, and Tat{-}Seng Chua. 2021.
\newblock \href {http://arxiv.org/abs/2101.00774} {Retrieving and reading: {A} comprehensive survey on open-domain question answering}.
\newblock \emph{arXiv preprint arXiv:2101.00774}.

\end{thebibliography}

\appendix

\clearpage

\section{Additional Experimental Setups}
\label{appendix:setup}
\begin{figure}
    \centering
    \includegraphics[width=0.95\linewidth]{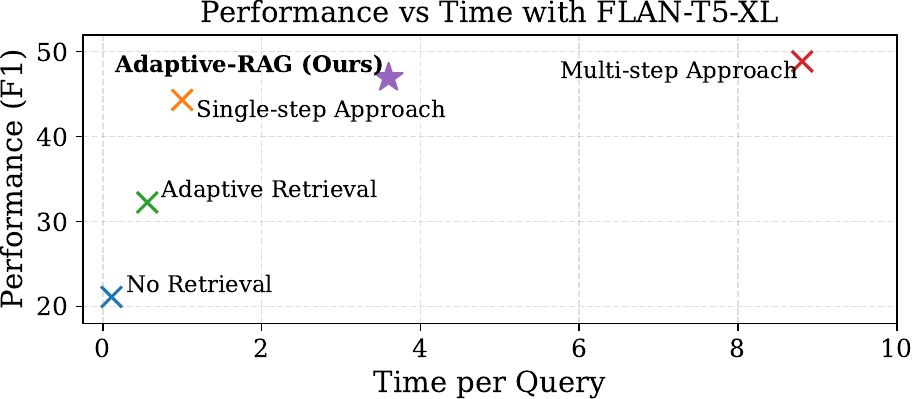}
    \vspace{-0.1in}
    \caption{QA performance (F1) and efficiency (Time/Query) for different retrieval-augmented generation approaches. We use the FLAN-T5-XL (3B) as the base LLM.}
    \label{fig:perf_effi_xl}
\end{figure}

\subsection{Datasets}
We use publicly open datasets for both single-hop and multi-hop QA datasets, referring to as~\citet{DPR} and \citet{ircot}, respectively. We describe the characteristics of each dataset:

\noindent
\textbf{1) SQuAD v1.1}~\cite{squad} is created through a process where annotators write questions based on the documents they read. 

\noindent
\textbf{2) Natural Questions}~\cite{nq} is constructed by real user queries on Google Search. 

\noindent
\textbf{3) TriviaQA}~\cite{tqa} comprises trivia questions sourced from various quiz websites.

\noindent
\textbf{4) MuSiQue}~\cite{musique} is collected by compositing multiple single-hop queries, to form queries spanning 2-4 hops. 

\noindent
\textbf{5) HotpotQA}~\cite{hotpotqa} is constructed by having annotators create questions that link multiple Wikipedia articles. 

\noindent
\textbf{6) 2WikiMultiHopQA}~\cite{2hop} is derived from Wikipedia and its associated knowledge graph path, needing 2-hops.


\subsection{Models}
We describe the details of models as follows:

\noindent
\textbf{1) No Retrieval.} This approach uses only the LLM itself, to generate the answer to the given query. 

\noindent
\textbf{2) Single-step Approach.} This approach first retrieves the relevant knowledge with the given query from the external knowledge sources and then augments the LLM with this retrieved knowledge to generate the answer, which iterates only once.

\noindent
\textbf{3) Adaptive Retrieval.} This baseline~\cite{adaptiveRetrieval} adaptively augments the LLM with the retrieval module, only when the entities appearing in queries are less popular. To extract entities, we use the available entity-linking method~\cite{blinke}, namely BLINK, for questions.

\noindent
\textbf{4) Self-RAG.} This baseline~\cite{self-rag} trains the LLM to adaptively perform retrieval and generation, where the retrieval is conducted once it predicts the special retrieval token above a certain threshold, and the answer generation follows.

\noindent
\textbf{5) Adaptive-RAG.} This is our model that adaptively selects the retrieval-augmented generation strategy, smoothly oscillating between the non-retrieval, single-step approach, and multi-step approaches\footnote{For the multi-step approach, we use the state-of-the-art question answering strategy from IRCoT~\cite{ircot}.} without architectural changes, based on the query complexity assessed by the classifier. 

\noindent
\textbf{6) Multi-step Approach.} This approach~\cite{ircot} is the multi-step retrieval-augmented LLM, which iteratively accesses both the retriever and LLM with interleaved Chain-of-Thought reasoning~\cite{cot} repeatedly until it derives the solution or reaches the maximum step number.

\noindent
\textbf{7) Adaptive-RAG w/ Oracle} This is an ideal scenario of our Adaptive-RAG equipped with an oracle classifier that perfectly categorizes the query complexity.

\subsection{Implementation Details}
For computing resources, we use A100 GPUs with 80GB memory. In addition, due to the significant costs associated with evaluating retrieval-augmented generation models, we perform experiments with a single run. Finally, we implemented models using PyTorch~\cite{DBLP:conf/nips/PaszkeGMLBCKLGA19} and Transformers library~\cite{DBLP:conf/emnlp/WolfDSCDMCRLFDS20}.

\section{Additional Experimental Results}

\begin{figure}
    \centering
    \includegraphics[width=0.95\linewidth]{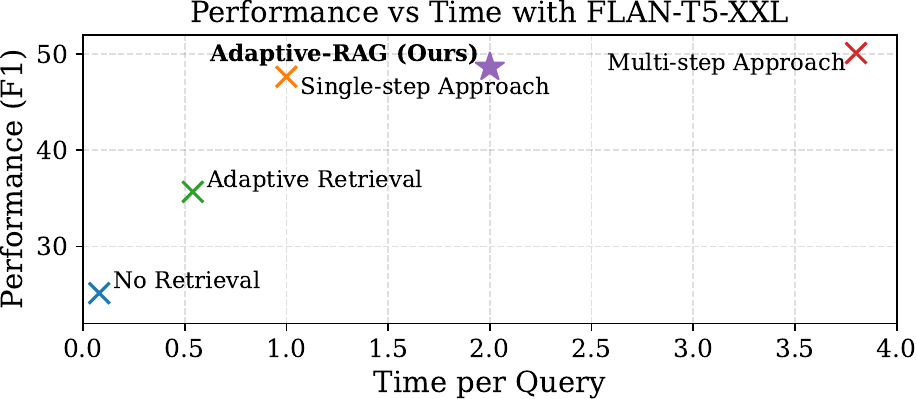}
    \vspace{-0.1in}
    \caption{QA performance (F1) and efficiency (Time/Query) for different retrieval-augmented generation approaches. We use the FLAN-T5-XXL (11B) as the base LLM.}
    \label{fig:perf_effi_xxl}
\end{figure}

\begin{table*}[t!]
\caption{Results on each of a collection of datasets with FLAN-T5-XXL (11B) as the LLM. We emphasize our results in bold.}
\vspace{-0.1in}
\label{tab:main:xxl}
\small
\centering
\resizebox{\textwidth}{!}{
\renewcommand{\arraystretch}{1.0}
\begin{tabular}{lllccccccccccccccc}
\toprule

& & & \multicolumn{5}{c}{\bf SQuAD} & \multicolumn{5}{c}{\bf Natural Questions} & \multicolumn{5}{c}{\bf TriviaQA} \\
\cmidrule(l{2pt}r{2pt}){4-8} \cmidrule(l{2pt}r{2pt}){9-13} \cmidrule(l{2pt}r{2pt}){14-18}
\textbf{Data} & \textbf{Types} &\textbf{Methods} & EM & F1 & Acc & Step & Time & EM & F1 & Acc & Step & Time & EM & F1 & Acc & Step & Time \\

\midrule

\multirowcell{7}[-0.0ex][l]{\textbf{Single-step}} 

& \multirowcell{2}[-0.0ex][l]{\textbf{Simple}} 

& \textbf{No Retrieval} & 7.00 & 14.40 & 8.40 & 0.00 & 0.08 & 18.80 & 25.50 & 20.40 & 0.00 & 0.08 & 32.80 & 39.20 & 35.40 & 0.00 & 0.08 \\

& & \textbf{Single-step Approach} & 28.80 & 40.80 & 35.00 & 1.00 & 1.00 & 41.40 & 51.20 & 47.60 & 1.00 & 1.00 & 56.00 & 64.70 & 61.80 & 1.00 & 1.00 \\

\noalign{\vskip 0.25ex}\cdashline{2-18}\noalign{\vskip 0.75ex}

& \multirowcell{3}[-0.0ex][l]{\textbf{Adaptive}} 

& \textbf{Adaptive Retrieval} & 15.60 & 25.60 & 20.00 & 0.50 & 0.54 & 31.00 & 39.70 & 35.00 & 0.50 & 0.54 & 44.80 & 52.20 & 48.60 & 0.50 & 0.54 \\

& & \textbf{Self-RAG$^*$} & 1.60 & 11.90 & 20.80 & 0.59 & 0.31 & 39.20 & 47.10 & 42.40 & 0.75 & 0.09 & 14.60 & 33.70 & 60.20 & 0.76 & 0.22 \\

& & \textbf{Adaptive-RAG (Ours)} & \textbf{27.80} & \textbf{39.80} & \textbf{34.00} & \textbf{1.17} & \textbf{1.50} & \textbf{41.20} & \textbf{51.00} & \textbf{47.40} & \textbf{1.00} & \textbf{1.00} & \textbf{52.00} & \textbf{60.30} & \textbf{57.20} & \textbf{1.03} & \textbf{1.33} \\

\noalign{\vskip 0.25ex}\cdashline{2-18}\noalign{\vskip 0.75ex}

& \textbf{Complex}

& \textbf{Multi-step Approach} & 24.60 & 36.90 & 30.20 & 2.13 & 3.83 & 39.60 & 49.60 & 46.40 & 2.16 & 3.94 & 52.60 & 61.10 & 59.40 & 2.17 & 4.03 \\

\noalign{\vskip 0.25ex}\cdashline{2-18}\noalign{\vskip 0.75ex}

& \textbf{Oracle}
& \textbf{Adaptive-RAG w/ Oracle} & 32.80 & 46.90 & 38.20 & 0.85 & 0.94 & 51.20 & 61.00 & 57.00 & 0.71 & 0.91 & 63.40 & 71.30 & 68.20 & 0.51 & 0.60 \\

\midrule
\midrule

& & & \multicolumn{5}{c}{\bf MuSiQue} & \multicolumn{5}{c}{\bf HotpotQA} & \multicolumn{5}{c}{\bf 2WikiMultiHopQA} \\
\cmidrule(l{2pt}r{2pt}){4-8} \cmidrule(l{2pt}r{2pt}){9-13} \cmidrule(l{2pt}r{2pt}){14-18}
\textbf{Data} & \textbf{Types} &\textbf{Methods} & EM & F1 & Acc & Step & Time & EM & F1 & Acc & Step & Time & EM & F1 & Acc & Step & Time \\

\midrule

\multirowcell{7}[-0.0ex][l]{\textbf{Multi-step}} 

& \multirowcell{2}[-0.0ex][l]{\textbf{Simple}} 

& \textbf{No Retrieval} & 4.20 & 13.40 & 5.40 & 0.00 & 0.08 & 17.40 & 25.44 & 18.40 & 0.00 & 0.09 & 26.80 & 32.93 & 28.00 & 0.00 & 0.08 \\

& & \textbf{Single-step Approach} & 16.80 & 25.70 & 19.20 & 1.00 & 1.00 & 37.60 & 49.27 & 39.60 & 1.00 & 1.00 & 46.60 & 54.13 & 48.20 & 1.00 & 1.00 \\

\noalign{\vskip 0.25ex}\cdashline{2-18}\noalign{\vskip 0.75ex}

& \multirowcell{3}[-0.0ex][l]{\textbf{Adaptive}} 

& \textbf{Adaptive Retrieval} & 8.40 & 17.80 & 10.20 & 0.50 & 0.54 & 26.60 & 36.01 & 27.80 & 0.50 & 0.54 & 35.20 & 42.68 & 36.80 & 0.50 & 0.54 \\

& & \textbf{Self-RAG$^*$} & 1.20 & 8.20 & 11.80 & 0.68 & 0.27 & 5.60 & 17.86 & 30.60 & 0.76 & 0.26 & 3.00 & 19.14 & 39.00 & 0.90 & 0.25 \\

& & \textbf{Adaptive-RAG (Ours)} & \textbf{20.60} & \textbf{28.50} & \textbf{23.20} & \textbf{1.89} & \textbf{3.12} & \textbf{44.20} & \textbf{54.78} & \textbf{46.80} & \textbf{1.58} & \textbf{2.53} & \textbf{47.60} & \textbf{57.36} & \textbf{54.00} & \textbf{1.46} & \textbf{2.55} \\

\noalign{\vskip 0.25ex}\cdashline{2-18}\noalign{\vskip 0.75ex}

& \textbf{Complex}

& \textbf{Multi-step Approach} & 19.40 & 27.50 & 21.80 & 2.09 & 3.66 & 47.00 & 57.81 & 49.40 & 2.08 & 3.73 & 57.60 & 67.65 & 64.00 & 2.17 & 3.63 \\

\noalign{\vskip 0.25ex}\cdashline{2-18}\noalign{\vskip 0.75ex}

& \textbf{Oracle}
& \textbf{Adaptive-RAG w/ Oracle} & 24.20 & 37.20 & 26.60 & 1.22 & 1.71 & 52.20 & 64.80 & 54.60 & 0.92 & 1.33 & 59.20 & 70.40 & 68.60 & 0.82 & 1.14 \\

\bottomrule

\end{tabular}
}
\end{table*}

\begin{table*}[t!]
\caption{Results on each of a collection of datasets with GPT-3.5 (Turbo) as the LLM. We emphasize our results in bold.}
\vspace{-0.1in}
\label{tab:main:gpt}
\small
\centering
\resizebox{\textwidth}{!}{
\renewcommand{\arraystretch}{1.0}
\begin{tabular}{lllccccccccccccccc}
\toprule

& & & \multicolumn{5}{c}{\bf SQuAD} & \multicolumn{5}{c}{\bf Natural Questions} & \multicolumn{5}{c}{\bf TriviaQA} \\
\cmidrule(l{2pt}r{2pt}){4-8} \cmidrule(l{2pt}r{2pt}){9-13} \cmidrule(l{2pt}r{2pt}){14-18}
\textbf{Data} & \textbf{Types} &\textbf{Methods} & EM & F1 & Acc & Step & Time & EM & F1 & Acc & Step & Time & EM & F1 & Acc & Step & Time \\

\midrule

\multirowcell{7}[-0.0ex][l]{\textbf{Single-step}} 

& \multirowcell{2}[-0.0ex][l]{\textbf{Simple}} 

& \textbf{No Retrieval} & 16.00 & 29.20 & 23.80 & 0.00 & 0.62 & 39.80 & 55.70 & 55.00 & 0.00 & 0.56 & 64.00 & 75.60 & 75.80 & 0.00 & 0.68 \\

& & \textbf{Single-step Approach} & 18.00 & 33.80 & 29.20 & 1.00 & 1.00 & 32.40 & 46.80 & 54.80 & 1.00 & 1.00 & 55.20 & 66.50 & 65.80 & 1.00 & 1.00 \\

\noalign{\vskip 0.25ex}\cdashline{2-18}\noalign{\vskip 0.75ex}

& \multirowcell{3}[-0.0ex][l]{\textbf{Adaptive}} 

& \textbf{Adaptive Retrieval} & 15.40 & 30.00 & 24.40 & 0.50 & 0.81 & 36.40 & 51.20 & 56.60 & 0.50 & 0.78 & 62.00 & 71.90 & 72.20 & 0.50 & 0.84 \\

& & \textbf{Self-RAG$^*$} & 1.60 & 11.90 & 20.80 & 0.59 & 1.91 & 39.20 & 47.10 & 42.40 & 0.75 & 0.52 & 14.60 & 33.70 & 60.20 & 0.76 & 1.59 \\

& & \textbf{Adaptive-RAG (Ours)} & \textbf{19.80} & \textbf{34.40} & \textbf{30.00} & \textbf{0.87} & \textbf{1.21} & \textbf{36.80} & \textbf{52.00} & \textbf{56.60} & \textbf{0.68} & \textbf{0.86} & \textbf{62.40} & \textbf{73.80} & \textbf{73.80} & \textbf{0.22} & \textbf{0.79} \\

\noalign{\vskip 0.25ex}\cdashline{2-18}\noalign{\vskip 0.75ex}

& \textbf{Complex}

& \textbf{Multi-step Approach} & 17.40 & 31.50 & 26.20 & 2.50 & 3.24 & 35.60 & 49.70 & 57.80 & 2.58 & 3.79 & 54.80 & 67.10 & 68.00 & 2.30 & 2.65 \\

\noalign{\vskip 0.25ex}\cdashline{2-18}\noalign{\vskip 0.75ex}

& \textbf{Oracle}
& \textbf{Adaptive-RAG w/ Oracle} & 28.00 & 45.90 & 39.40 & 0.54 & 0.93 & 50.00 & 65.40 & 67.00 & 0.28 & 0.8 & 70.80 & 81.00 & 80.00 & 0.11 & 0.73 \\

\midrule
\midrule

& & & \multicolumn{5}{c}{\bf MuSiQue} & \multicolumn{5}{c}{\bf HotpotQA} & \multicolumn{5}{c}{\bf 2WikiMultiHopQA} \\
\cmidrule(l{2pt}r{2pt}){4-8} \cmidrule(l{2pt}r{2pt}){9-13} \cmidrule(l{2pt}r{2pt}){14-18}
\textbf{Data} & \textbf{Types} &\textbf{Methods} & EM & F1 & Acc & Step & Time & EM & F1 & Acc & Step & Time & EM & F1 & Acc & Step & Time \\

\midrule

\multirowcell{7}[-0.0ex][l]{\textbf{Multi-step}} 

& \multirowcell{2}[-0.0ex][l]{\textbf{Simple}} 

& \textbf{No Retrieval} & 20.40 & 31.30 & 24.40 & 0.00 & 0.81 & 37.40 & 51.04 & 43.20 & 0.00 & 0.74 & 37.00 & 48.50 & 43.40 & 0.00 & 0.90 \\

& & \textbf{Single-step Approach} & 16.40 & 26.70 & 23.60 & 1.00 & 1.00 & 39.60 & 50.44 & 45.60 & 1.00 & 1.00 & 46.80 & 57.69 & 52.60 & 1.00 & 1.00 \\

\noalign{\vskip 0.25ex}\cdashline{2-18}\noalign{\vskip 0.75ex}

& \multirowcell{3}[-0.0ex][l]{\textbf{Adaptive}} 

& \textbf{Adaptive Retrieval} & 18.80 & 30.30 & 24.80 & 0.50 & 0.90 & 38.60 & 50.70 & 43.20 & 0.50 & 0.87 & 44.20 & 55.11 & 50.60 & 0.50 & 0.95 \\

& & \textbf{Self-RAG$^*$} & 1.20 & 8.20 & 11.80 & 0.68 & 1.66 & 5.60 & 17.86 & 30.60 & 0.76 & 1.67 & 3.00 & 19.14 & 39.00 & 0.90 & 1.81 \\

& & \textbf{Adaptive-RAG (Ours)} & \textbf{21.80} & \textbf{32.60} & \textbf{29.60} & \textbf{1.90} & \textbf{2.29} & \textbf{40.40} & \textbf{52.56} & \textbf{47.00} & \textbf{0.93} & \textbf{1.48} & \textbf{46.60} & \textbf{60.09} & \textbf{56.80} & \textbf{1.59} & \textbf{2.23} \\

\noalign{\vskip 0.25ex}\cdashline{2-18}\noalign{\vskip 0.75ex}

& \textbf{Complex}

& \textbf{Multi-step Approach} & 23.00 & 32.50 & 31.60 & 3.41 & 3.61 & 45.80 & 58.36 & 52.20 & 2.73 & 3.18 & 52.20 & 66.08 & 62.40 & 3.36 & 3.35 \\

\noalign{\vskip 0.25ex}\cdashline{2-18}\noalign{\vskip 0.75ex}

& \textbf{Oracle}
& \textbf{Adaptive-RAG w/ Oracle} & 29.60 & 44.70 & 35.60 & 0.90 & 1.45 & 55.60 & 69.90 & 62.80 & 0.54 & 1.08 & 52.20 & 69.90 & 66.60 & 0.65 & 1.21 \\

\bottomrule

\end{tabular}
}
\end{table*}

\paragraph{Performance vs Time}
We further provide a comparison of different retrieval-augmented generation approaches with FLAN-T5-XL and FLAN-T5-XXL models in Figure~\ref{fig:perf_effi_xl} and Figure~\ref{fig:perf_effi_xxl}, respectively, in the context of performance and efficiency trade-offs. Similar to the observation made from the GPT-3.5 model in Figure~\ref{fig:perf_effi_gpt}, our proposed Adaptive-RAG is significantly more effective as well as efficient.

\paragraph{Performance per Dataset}
In addition to detailing the performance of each dataset with the FLAN-T5-XL model, as shown in Table \ref{tab:main:xl}, we also present the results for each dataset with the FLAN-T5-XXL and GPT-3.5 models in Table~\ref{tab:main:xl} and Table~\ref{tab:main:gpt}, respectively. The experimental results show that our Adaptive-RAG consistently balances between efficiency and accuracy. It is worth noting that while the GPT-3.5 model performs effectively in addressing straightforward queries even without document retrieval, it benefits significantly from our Adaptive-RAG in terms of effectiveness when solving complex multi-hop queries.

\end{document}